\documentclass{article}
\usepackage{amssymb}

\usepackage{multicol}
\usepackage{multirow}
\usepackage[bookmarks=true]{hyperref}
\usepackage{xcolor}
\usepackage{xspace}
\usepackage{graphicx}
\usepackage{siunitx}
\usepackage{caption}
\usepackage{subcaption}
\usepackage[dvipsnames]{xcolor}
\usepackage{amssymb, amsfonts, amsmath}
\usepackage{makecell}
\usepackage{booktabs}
\usepackage{wrapfig}
\usepackage{subcaption}
\usepackage{enumitem}
\usepackage{titlesec}

\usepackage[final]{corl_2025} %

\title{DexSkin: High-Coverage Conformable Robotic Skin for Learning Contact-Rich Manipulation}

\def\name{\mbox{DexSkin}\xspace}
\newcommand{\eqcmark}[1]{\footnotemark[1]\kern0.4em\textsuperscript{#1}}

\author{
\textbf{Suzannah Wistreich\thanks{indicates equal contribution}\kern0.4em\textsuperscript{1}
, Baiyu Shi\eqcmark{1} , Stephen Tian\eqcmark{1},} \\ \textbf{Samuel Clarke\textsuperscript{1}, Michael Nath\textsuperscript{1}, Chengyi Xu\textsuperscript{1,2}, Zhenan Bao\textsuperscript{1}, Jiajun Wu\textsuperscript{1}}\\
\textsuperscript{1}\textnormal{Stanford University} \hspace{0.05cm}
\textsuperscript{2}\textnormal{University of Alabama at Birmingham}
}

\begin{document}
\maketitle

\vspace{-1em}

\begin{abstract}
Human skin provides a rich tactile sensing stream, localizing intentional and unintentional contact events over a large and contoured region. 
Replicating these tactile sensing capabilities for dexterous robotic manipulation systems remains a longstanding challenge.
In this work, we take a step towards this goal by introducing \name. \name is a soft, conformable capacitive electronic skin that enables sensitive, localized, and calibratable tactile sensing, and can be tailored to varying geometries.
We demonstrate its efficacy for learning downstream robotic manipulation by sensorizing a pair of parallel jaw gripper fingers, providing tactile coverage across almost the entire finger surfaces.
We empirically evaluate \name's capabilities in learning challenging manipulation tasks that require sensing coverage across the entire surface of the fingers, such as reorienting objects in hand and wrapping elastic bands around boxes, in a learning-from-demonstration framework. We then show that, critically for data-driven approaches, \name can be calibrated to enable model transfer across sensor instances, and demonstrate its applicability to online reinforcement learning on real robots. 
Our results highlight \name's suitability and practicality for learning real-world, contact-rich manipulation.
Please see our project webpage for videos and visualizations: \url{https://dex-skin.github.io/}.

\end{abstract}
\keywords{tactile sensing, contact-rich manipulation} 

\begin{center}
    \includegraphics[width=\textwidth]{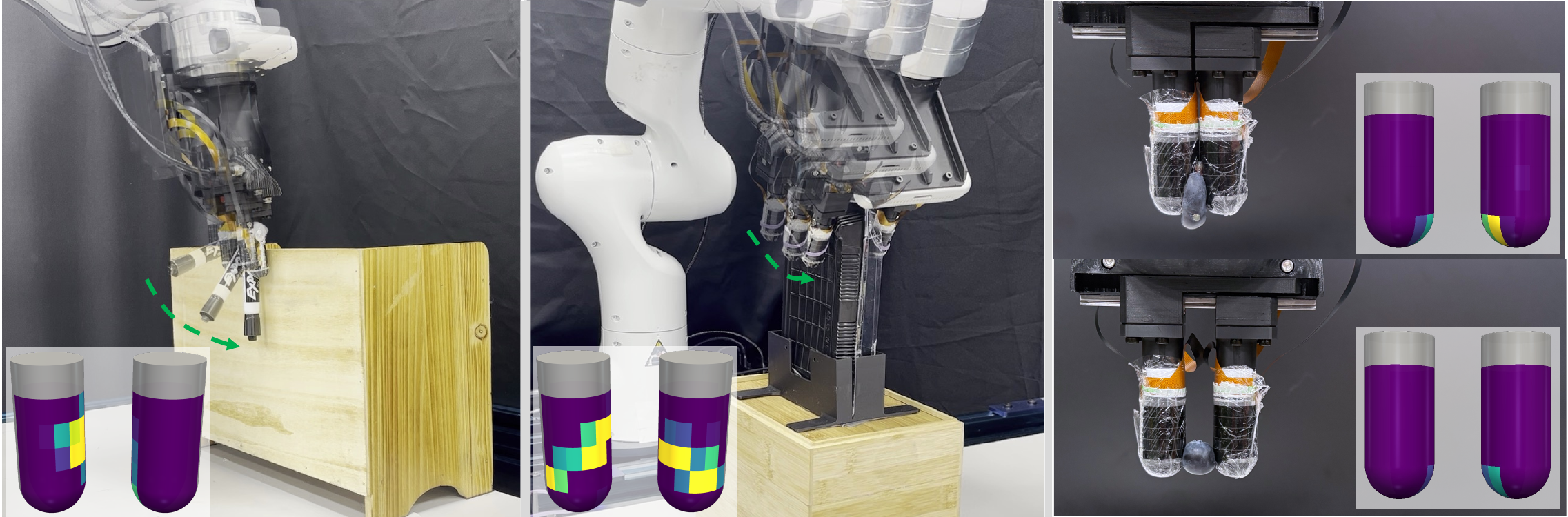}
    \captionof{figure}{ \small
    \textbf{\name sensors enable contact-rich robotic manipulation policies.} We demonstrate the applications of \name in in-hand pen reorientation, box packaging, and berry transporting tasks. In each frame, we visualize tactile readings from \name as heatmaps, where brighter readings correspond to larger forces. \name's high-coverage sensing capabilities enable it to provide contact signals for a variety of robotic manipulation tasks. At the same time, it possesses desirable properties for learning-based manipulation systems.}
    \label{fig:teaser}
    \end{center}

\section{Introduction}
Tactile feedback is essential for robust and dexterous manipulation in natural and artificial systems. In humans, mechanoreceptors within the skin provide a rich sensory stream that guides tasks ranging from handling delicate objects to using tools with force~\cite{johansson2009coding}. This tactile feedback enables more precise reactive control than can be achieved with human and proprioceptive feedback alone~\cite{macefield1996control}. 

Emulating this tactile sensitivity in robotic systems has long been a challenge, starting from sensing hardware. Research efforts typically focus on rigid or partially flexible sensors that offer low spatial coverage, limited adaptability, and poor conformability to complex surfaces. In contrast, everyday tasks such as rotating a key in-hand or picking up delicate berries require sensing coverage in multiple contact regions. 
Similarly, wrapping an elastic band around an object requires a sensor to detect, distinguish, and localize dynamic contact events on all surfaces of a human hand or end effector accurately, applying the appropriate force without allowing the band to slip.

Learning-based approaches offer a general way to harness tactile sensing information. However, data-driven systems pose requirements beyond sensing coverage. Seemingly small distribution shifts in sensor readings from wear or replacement can render previously trained models unusable, thus, output signals must be replicable across sensor hardware instances. Additionally, for real-world online learning, sensors must remain consistent, durable, and precise under repeated stresses during trial-and-error interaction, and output interpretable signals amenable to reward or cost specification.

To address these challenges, we introduce a novel soft tactile skin named \textbf{\name} that is particularly suitable for robotic learning applications. \name can be conformably integrated onto robotic end-effectors with unparalleled spatial coverage. 
It is based on a capacitive mechanism, and features high sensitivity and robustness under repeated interactions. 
Because each of the dozens of taxels on the skin are individually addressable, \name can localize and characterize simultaneous contacts from distinct regions.
It can also be calibrated to provide consistent readings across distinct sensor units, enabling re-use of learned networks. At the same time, it can withstand deformations encountered during typical dexterous tasks such as pinching, twisting, and bending.

In this work, we introduce the \name framework and its fabrication, as well as a representative integration with a soft cylindrical robotic fingertip that sensorizes the distal dome of the finger and 294$^{\circ}$ of the circumference. Then, we evaluate \name's applicability to robot learning. First, we test whether \name's coverage and tailorability expands the range of learnable manipulation tasks. Then, we evaluate its calibration performance, which are critical for working with and transferring learned tactile models. Finally, we demonstrate its suitability for online robotic learning settings by performing real-world reinforcement learning for a delicate object picking task. The results highlight \name's robust applicability to a wide range of robotic tasks and morphologies, and its particular practicality for robot learning researchers and practitioners.

\section{Related Work}

\textbf{Tactile sensing in robotics.}
Tactile sensors generally fall into three categories: vision-based and optical, magnetic, and electrically addressable.

Vision-based sensors image high-resolution deformations of an elastomer surface using cameras. 
Various designs have been developed for flat~\cite{johnson2011microgeometry, li2014localization,donlon2018gelslim, lambeta2020digit}
or round and dome-shaped surfaces~\cite{alspach2019soft, gomes2020geltip, kuppuswamy2020soft, do2022densetact, do2023densetact, do2024densetact, lambeta2024digitizing}.
Miniaturizing and customizing them has proven challenging~\cite{do2024densetact,di2024using}, and each target form factor requires re-design with extensive optical considerations. Alternatively, DISCO~\cite{piacenza2020sensorized} uses photodiodes and LEDs beneath a reflective elastomer, but requires an extensive calibration process to correlate signals with contact locations and magnitudes.

Magnetic-based sensors~\cite{uskin2018, bhirangi2021reskin, bhirangi2024anyskin}
measure changes in magnetic flux of a magnetized surface due to deformation. Assembling rigid magnetometers onto commercial boards requires lengthy design iterations and is challenging to fit to complex geometries. Further, the sparse point readings struggle to resolve contact imprints or simultaneous contacts, and estimating contact locations and magnitudes demands a data-intensive learning process. In contrast, \name is intrinsically stretchable and conformable to complex geometries. Its individually addressable taxels resolve multiple contacts. Sensors from different batches can be aligned in minutes and calibrating for normal force measurements requires just four loading-unloading cycles per taxel.

Electrically addressable tactile sensors harness resistive, impedance, and capacitive mechanisms. 
Piezoresistive sensors~\cite{Resistive2021Ding, luo2021learning, sundaram2019STAG} detect force via pressure-sensitive layers but often have low sensitivity, large hysteresis, and temperature dependence. AC-driven impedance-based sensors (e.g., BioTac~\cite{Biotac2008}) tend to be costly, low-density, and limited in customization. Capacitive sensors~\cite{Cap2020Bai, Cap2021Luo, Cap2019Yang, Cap2020Huang} measure displacement between electrode plates to resolve force, offering high sensitivity and low power consumption, but often require  microfabrication (e.g., photolithography).
\name leverages the capacitive mechanism for performance, but can be fabricated on-site via accessible processes, enabling cost-effective same-day customization of sensing pattern and geometry.

\textbf{Learning robotic manipulation with tactile sensors.}
Many prior works use tactile sensors toward learning robotic manipulation. Some train in simulation and later transfer to the real world~\cite{qi2023general, suresh2023neural, yuan2023robot, xu2024geotact}. %
Since accurately simulating contacts and sensor outputs is difficult, many instead learn directly from real-world data, for instance performing model-free reinforcement learning (RL)~\cite{lee2019making, dong2021tactile, kim2022active, yu2024mimictouch}, learning supervised or self-supervised predictive models~\cite{calandra2017feeling, calandra2018more, tian2019manipulation, lambeta2020digit}, or imitating expert demonstrations~\cite{li2023see, yu2024mimictouch, huang20243d, lin2024learning}. Real-world tactile data can be fused with other observation modalities~\cite{zhao2025polytouchrobustmultimodaltactile} or provide rapid feedback~\cite{xue2025reactive}. We operate in this real-world data paradigm. 
While we do not introduce new learning algorithms in this work, our findings indicate that \name can expand the capabilities and practicality of a variety of learned tactile manipulation systems. 
Unlike prior works that perform imitation learning that demonstrate prehensile grasping~\cite{huang20243d}, packing~\cite{yu2024mimictouch}, and pouring~\cite{li2023see} tasks, we focus on tasks that require high-coverage sensing capabilities with minimal blind spots, support for multiple simultaneous contacts, and recovery from perturbation such as in-hand reorientation and manipulating elastic bands. 
Further, we test the transfer of learned models across \name sensors, a critical practical challenge. \citet{bhirangi2024anyskin} narrow inter-sensor gaps with sophisticated manufacturing techniques and hardware design. We instead perform rapid standardization to obtain mappings between sensor taxels, for localized, transferable sensing. 
In real-world online RL settings, while prior work mostly uses tactile information as policy inputs to optimize task completion objectives~\cite{lee2019making, dong2021tactile, kim2022active, yu2024mimictouch}, \name's readouts are more interpretable compared to optical or magnetic sensors. This allows rewards to be computed directly regarding the contact information itself, for instance, encouraging the gentle grasp of a berry.

\begin{wrapfigure}[18]{t}{0.48\textwidth}
    \vspace{-5em}
    \centering \includegraphics[width=\linewidth]{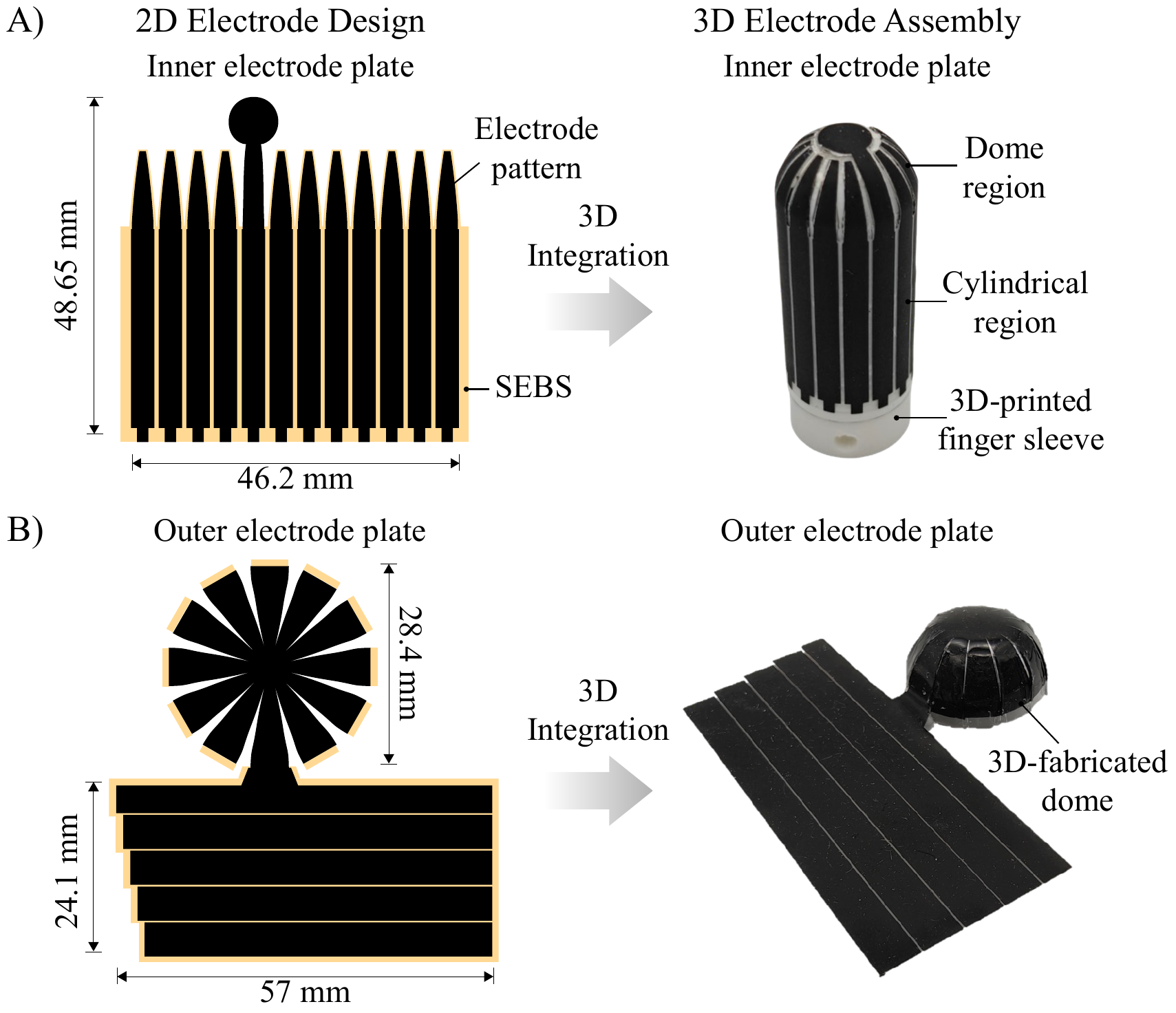}
    \caption{ \small \textbf{Designs of electrode patterns.} The 2D pattern design of the inner plate (top left) and outer plate (bottom left) electrodes on the soft SEBS substrate and its appearance after its vertical traces have been conformed onto the dome and cylindrical circumference of the finger sleeve (top right and bottom right). Note that SEBS is transparent and can be hard to distinguish in the post-assembly photo (right). 
    }
    \label{fig:electrodes}
\end{wrapfigure}

\section{The \name Framework}

In this section, we introduce the main components of the \name framework and compare it to other widely adopted tactile sensors in Table~\ref{tab:sensor_compare}. Although \name can be applied to a variety of morphologies, as shown in Appendix Figure~\ref{fig:more_morphologies}, we focus on a system that enables a robotic arm to learn tactile-based manipulation policies by sensorizing the fingertip-shaped jaws of a parallel gripper. We first discuss the soft sensor skin itself, its fabrication, then describe its integration with a finger, robotic gripper, and 7-axis arm. 

\textbf{Low-cost, high-performance, conformable \name sensors.}
\label{sec:sensor_fab}
Electrically addressable tactile sensors remain rare in robotics~\cite{huang20243d, huang2019learning} due to challenges in sensitivity, hysteresis, cost, and integration. We introduce \name, a soft stretchable capacitive sheet that can be fabricated for under \$10 per pair (at 1,000 units) with easily accessible fabrication tools while maintaining high sensitivity, commercial off-the-shelf cyclic drift and hysteresis performance (TekScan WB201), and a broad pressure sensing range (up to 702kPa). It provides almost full coverage for a robotic finger while offering calibratable and individually addressable taxels that inherently resolve simultaneous contacts. 
\begin{wraptable}{r}{0.49\textwidth}
\setlength{\tabcolsep}{5pt} %
\renewcommand{\arraystretch}{1.05} %
\small %
\begin{tabular}{@{}lrrr@{}}
\toprule
Sensor & \makecell{Area\\(mm²) $\uparrow$} & \makecell{Force res.\\(N) $\downarrow$} & \makecell{Spatial res. \\ (mm) $\downarrow$} \\
\midrule
\name\ (ours) & \makecell[r]{1944--6471+} & 0.086{\dag}       & $\leq0.60^*$ \\
ReSkin~\cite{bhirangi2021reskin}        & 400+          & $<0.2$     & 1.00       \\
BioTac~\cite{Biotac2008}        & 484           & 0.26      & 1.40       \\
DISCO~\cite{piacenza2020sensorized}         & 6107          & $<0.35$    & $\geq0.7$  \\
GelTip~\cite{gomes2020geltip}        & 3142          & N/A         & $\geq3.33$ \\
DIGIT~\cite{lambeta2020digit}         & 304           & 0.006      & 0.150      \\
DIGIT360~\cite{lambeta2024digitizing}      & 2340          & 0.001      & 0.007      \\
\bottomrule
\end{tabular}
  \caption{\small Comparison of DexSkin framework to state-of-the-art tactile sensors. \name enables the largest sensing area coverage, the finest spatial and force resolution among selected non-vision based systems. \dag RMSE of force estimation over $\SI{0} - \SI{2.5}{\newton}$ normal loads. *Estimated from pitch size, machine resolution dependent. More details are in Appendix~\ref{sec:characterization_experiments}.}
  \label{tab:sensor_compare}
\vspace{-1em}
  
\end{wraptable}

\name adopts a compliant parallel plate grid configuration where a highly deformable dielectric layer is sandwiched by two soft electrode plates, forming individually addressable capacitive sensing taxels at every electrode intersection.
We tailor the sensor to envelop a fingertip-like shape despite being a single continuous piece. It conforms to the curved sides of the sleeve and also to shapes with non-zero Gaussian curvature that challenge traditional sheet-like taxel-based sensors. The hemispherical dome of the finger adopts a flower petal-inspired design as the outer plate, as shown in Figure~\ref{fig:electrodes}(b). It uses protruding columns, which narrow toward the tip, as the inner plate. When the petals' edges are sealed, it becomes a snugly fitting hemispherical outer plate that forms 12 taxels with each column underneath. The cylindrical body features 48 taxels, located at the intersections of the black electrode crossbar structure formed by wrapping the row-wise outer plate around the column-wise inner plate (see  Figure~\ref{fig:electrodes}(a)) providing an angular coverage of 294$^{\circ}$ around the finger.

\begin{wrapfigure}[15]{rt}{0.49\textwidth}
    \vspace{-1em}
    \centering \includegraphics[width=\linewidth]{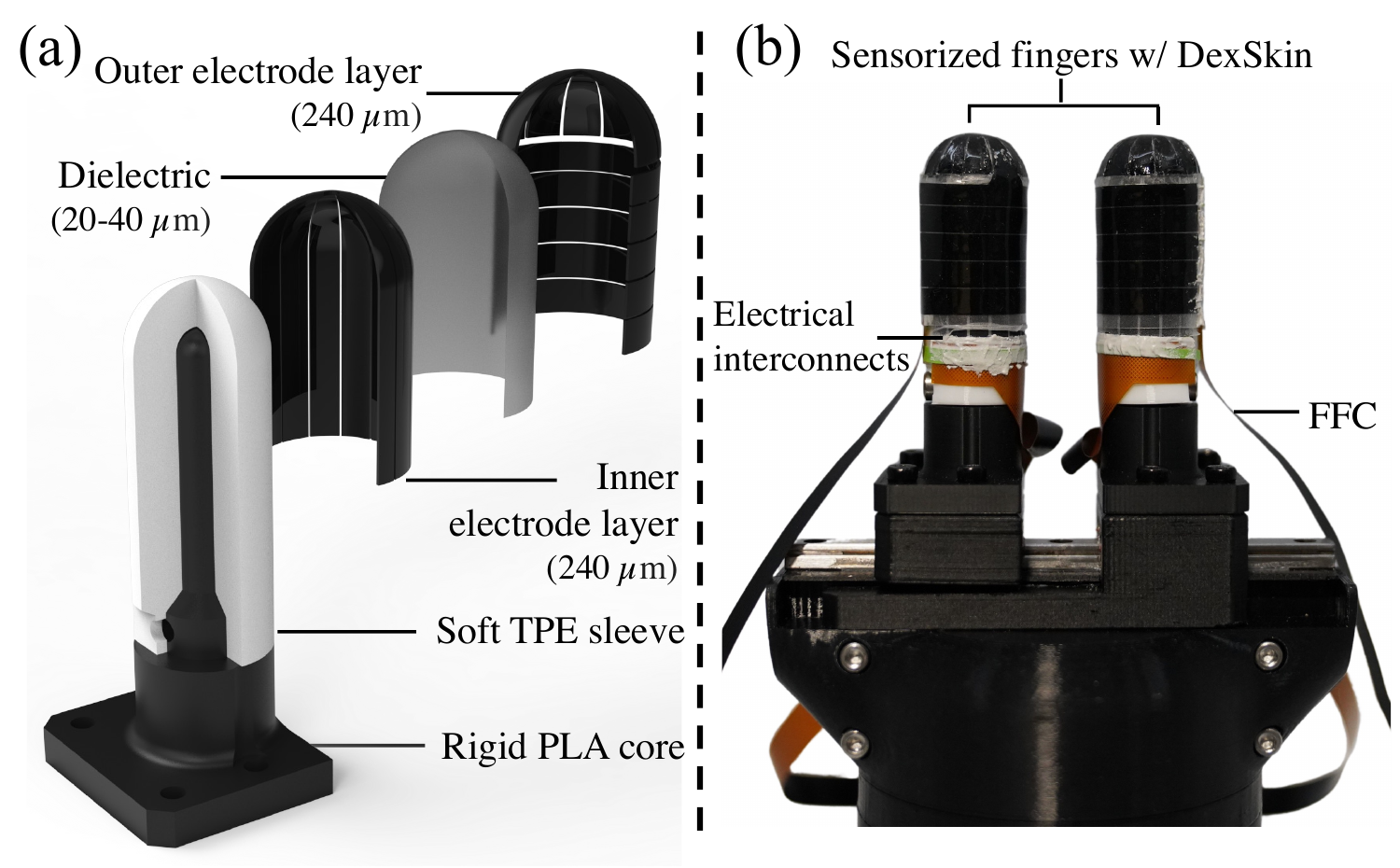}
    \caption{\small
    \textbf{Robotic gripper finger assembly sensorized with \name.} (a) Exploded section-view diagram of the finger assembly. 
    (b) Two fully assembled sensorized gripper fingers on the SSG-48 gripper used in our experiments.
    }
    \label{fig:gripper_design}
\end{wrapfigure}
\label{sec:experiment_setup}

\textbf{Same-day tailorable sensor fabrication.}
The fabrication process begins with computer-aided electrode pattern design, providing full control over taxel layout and size. 
We incorporate conductive wire traces and capacitive plate patterns into a styrene-ethylene-butylene-styrene (SEBS) substrate, resulting in a 240\textmu m-thick thin film.
We seal the edges of the flower-petal outer plate using drips of SEBS solution, to form the hemispherical structure shown in Figure~\ref{fig:gripper_design}(b). 
We screen-print silver paste adhesive (MG Chemicals 8331) through paper shadow masks to connect sensor traces to FFCs. For the dielectrics, we spincoat the SEBS solution onto a reusable micro-structured PDMS mold (Dow Sylgard\texttrademark \space 184) to produce the 20\textmu m-thick dielectric layer.

We assemble the sensor by first attaching the inner plate to the soft sleeve using double-sided tape. We then wrap the dielectric around the inner plate and cap it on top with the hemispherical plate. 

We conducted experiments characterizing \name's sensing range, crosstalk, cross-taxel uniformity, robustness, hysteresis, and cyclic stability. Please see Appendix~\ref{sec:characterization_experiments} for the details and results.
\textit{We are committed to open-sourcing the detailed fabrication instructions and materials for \name.}

\textbf{Integrated Robotic Setup.}
\label{sec:integrated_setup}
We integrate two \name-sensorized fingers into a tabletop manipulation system with a 7-axis Franka Emika Panda arm, shown in Appendix Figure~\ref{fig:robot_setup}.

While rigid gripper fingers may offer better sensor performance by causing most of the force-induced deformation to occur between its electrode plates, soft materials provide better cushioning for object interactions. 
Balancing compliance and sensor performance, we design each finger with three layers (Figure~\ref{fig:gripper_design}): the sensor skin, a deformable outer sleeve, and a rigid inner core for structural integrity. We 3D-print the sleeve and inner core using NinjaFlex TPU and PLA respectively.

For precise and reactive finger control, we mount the \name sensors to a Source Robotics SSG-48 gripper. A complete fabricated gripper assembly is shown in Figure~\ref{fig:gripper_design}(b). For tasks that use visual input, we mount a RealSense D415 RGB-D camera to the robot end-effector's wrist.

\begin{figure*}[t]
    \centering
    \vspace{-0.5em}
    \includegraphics[width=\textwidth]{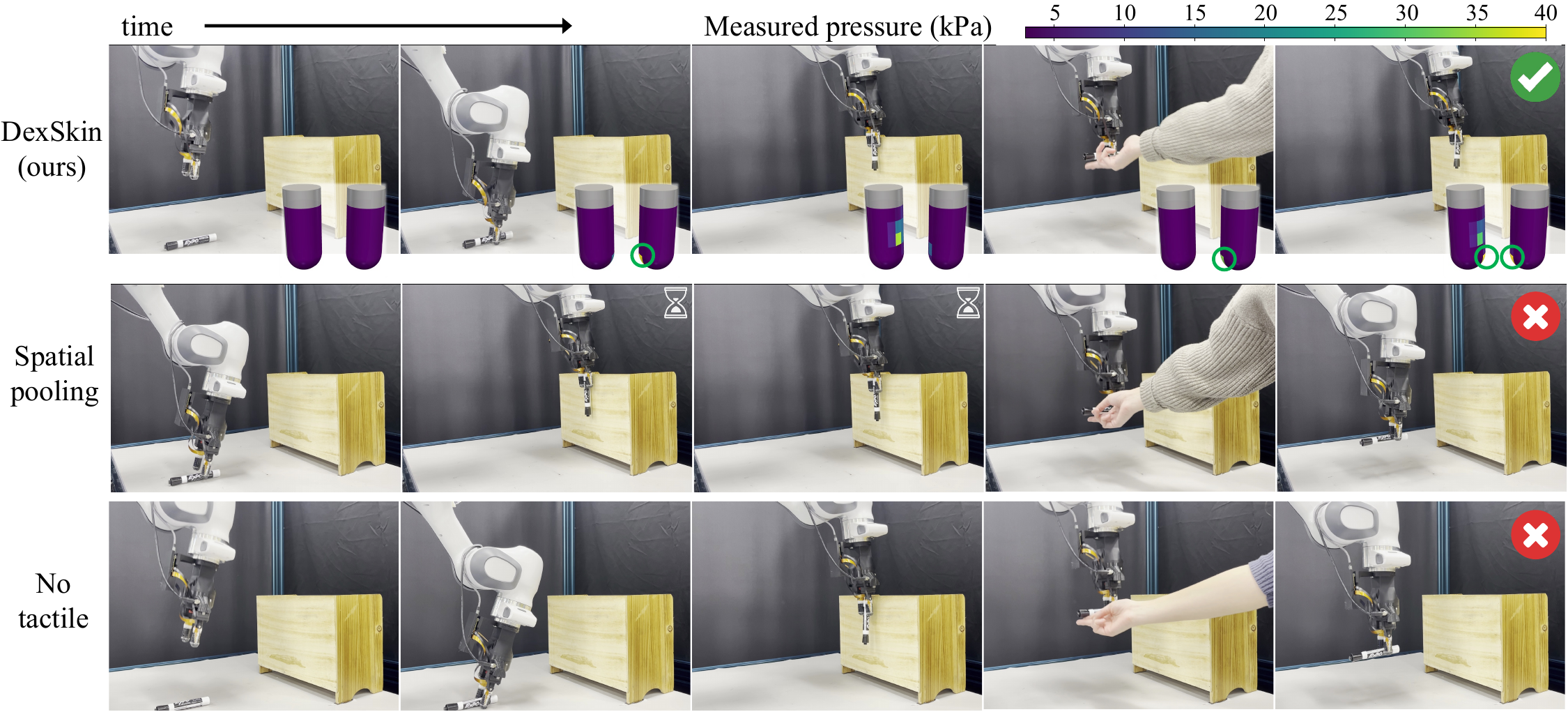}
    \vspace{-1.5em}
    \caption{\small \label{fig:pen_qualitative} Pen reorientation task rollouts. \textbf{\textit{(Top)}} \textbf{The \name policy successfully reorients the pen (cols. 1-3) and reacts to human perturbation (col. 4) by repeating the motion (col. 5).} We visualize \name readings as if facing the robot's front. Note the readings at the sensor's domed tip when the pen is grasped (\textcolor{ForestGreen}{green} circle, col. 2) and the impression when the pen is reoriented to vertical (cols. 3 \& 5). \textbf{\textit{(Middle)}} \textbf{With spatially pooled tactile information, the robot fails to respond to perturbation and stalls excessively (cols. 2-3)}. \textbf{\textit{(Bottom)}} \textbf{The no tactile policy is unable to detect perturbation}. 
    \textit{Best viewed in supplementary video.} 
    }
    \vspace{-1em} 
\end{figure*}

\vspace{-0.5em}
\section{Evaluating \name for Learning Robotic Manipulation}
\vspace{-0.5em}
We evaluate \name for learning robotic manipulation from the following perspectives:

\begin{enumerate}[leftmargin=*,topsep=0pt,partopsep=0pt,parsep=0pt,itemsep=2pt]
\item Can \name's coverage and tailorability enable robots to learn a range of manipulation tasks?
\item Can calibrating \name allow learned policies to be transferred across sensor instances?
\item Is \name suitable for applications 
 to learning robot behaviors \textit{online}?
\end{enumerate}
\vspace{-0.5em}
\subsection{Learning Manipulation with Expanded Coverage and Tailorability}
\vspace{-0.5em}
To test whether the coverage and resolution of \name can expand the manipulation capabilities of learned robotic policies, we consider two representative tasks. For each task, we learn policies via imitation learning from $50$ expert human demonstrations by teleoperating the setup from Section~\ref{sec:integrated_setup} using GELLO~\cite{wu2024gello} and training diffusion policies~\cite{chi2023diffusionpolicy, chi2024diffusionpolicy}.  
The tasks are as follows:
\begin{itemize}[leftmargin=*,topsep=0pt,partopsep=0pt,parsep=0pt,itemsep=2pt]
\item \textbf{In-hand pen reorientation.} The robot must pick up the pen and push it against a flat face of a nearby cabinet to reorient it in-hand, using only tactile and proprioceptive feedback. Further, it must be robust to disturbances (e.g. human perturbation) of the pen at inference time. This task requires the robot to use tactile information to estimate the in-hand pose of the pen.
\item \textbf{Packaging a box with a suitable elastic band.} Humans use the dorsal side of their fingertips to perform everyday tasks like flipping switches or removing batteries from remotes. To evaluate similar capabilities, we task the robot with securing the lid of a container with a rubber band. The robot is randomly provided with an intact band or a perforated band that will snap if used. In the latter case, the robot must retrieve a replacement. This is challenging because the robot must adjust its behavior based on forces exerted by the unknown band on the dorsal sides of the fingers.
\end{itemize}

\begin{figure*}[t]
    \centering
  \includegraphics[width=\textwidth]{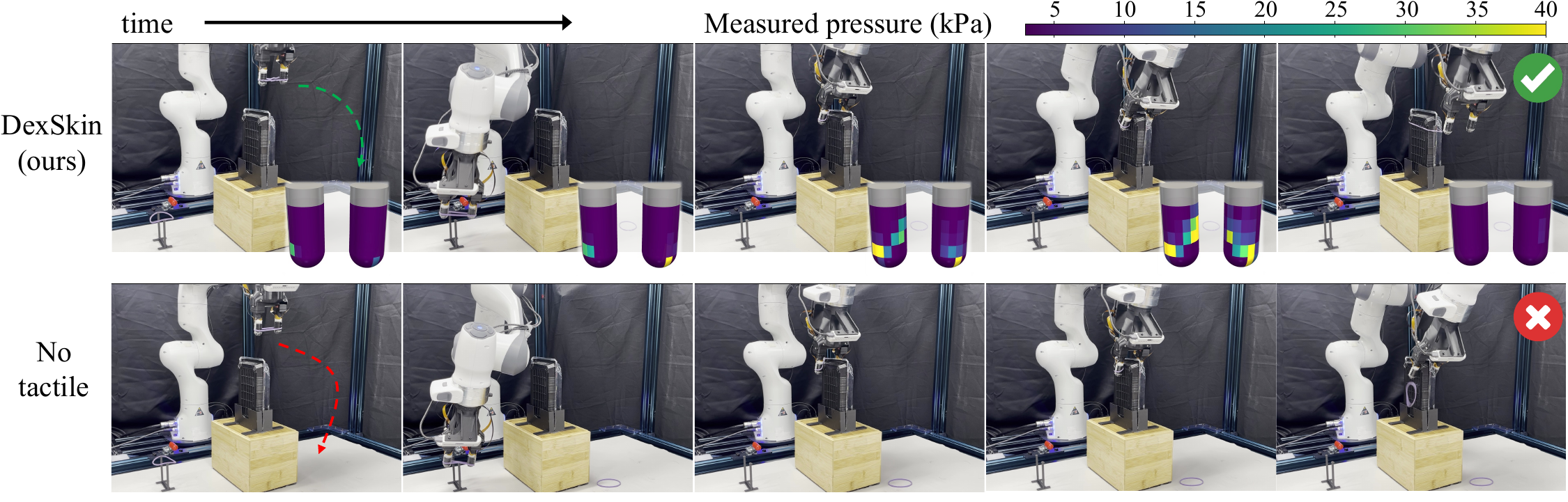}
    \caption{\small \label{fig:box_qualitative} Examples of box packaging task rollouts. \textbf{\textit{(Top)}} \textbf{With an initially provided perforated band, the \name policy correctly detects and discards the perforated band based on the tactile reading (col. 1), retrieves the backup (col. 2), and wraps it around the box (cols. 3-5).} We visualize \name sensor readings as if the viewer is facing the robot, looking at the gripper. \textbf{\textit{(Bottom)}} \textbf{With an initially provided \textit{un}perforated band, the policy without tactile information unnecessarily replaces the band (cols. 1-2) and then fails to package the box as the band slips off the gripper fingers (cols. 3-5).} 
    \textit{Best viewed in supplementary video.} 
    }
    \vspace{-2em}
\end{figure*}

\textbf{Comparative evaluation.}  To test our hypothesis, we compare the following settings:

\begin{itemize}[leftmargin=*,topsep=0pt,partopsep=0pt,parsep=0pt,itemsep=2pt]
\item \textit{\name (ours)}: Uses the full $120$ taxels of \name readings, proprioception, and wrist camera RGB images. Note wrist camera images are provided for the box task only, not the pen task.
\item \textit{No tactile}: A baseline identical to \textit{\name (ours)} but excluding all tactile information.
\item Ablated policies trained with spatially aggregated tactile readings for the pen task (\textit{\name (spatial pooling)}), emulating force-sensitive resistors or load cells that cannot detect multiple localized contacts, and a separate ablation using readings from only inner sensor columns between the gripper fingers for the box task (\textit{\name (inner cols. only)}), a proxy for sensors with flat or restricted sensing surfaces. These test the importance of \name's resolution and coverage.
\item \textit{DIGIT}: Uses a pair of DIGIT~\cite{lambeta2020digit} sensors as gripper fingers and as input to the policy, comparing our system to existing and commercially available sensors.
\end{itemize}
\label{sec:pen_task}

\textbf{Pen task results.} We test two settings: the first is the same as in training; in the second, if the robot reorients the pen successfully, a human experimenter perturbs the pen, rotating it within the robot's gripper to be again parallel to the ground. 
The results are shown in Table~\ref{tab:pen_box_packaging_results} and Figure~\ref{fig:pen_qualitative}. Without perturbation, three out of four policies solve the task $19/20$ times. 
The \name (spatial pooling) policy achieves moderate but poorer performance. When perturbed, the policy with full \name information maintains its success rate, and is the only policy that can recover from perturbation. 

Qualitatively, we find that the policy without tactile information repeats the same trajectory regardless of external perturbation, per expectation, making it effective only in the unperturbed setting. While one may anticipate that the policy with spatially pooled tactile information may struggle to estimate in-hand object pose and deal with perturbations, it does not succeed reliably in the unperturbed setting either, likely because it sometimes has insufficient information to determine whether the pen has already been reoriented or not. In its failure cases, this policy often does not fully reorient the pen. For DIGIT, while the sensor provides rich signal within its sensitive region, we find surprisingly that many contacts with the pen are outside this narrow region, even though demonstrations include natural motions with approximately antipodal grasps.

\textbf{Box packaging results.}
The results are shown in Table~\ref{tab:pen_box_packaging_results} and Figure~\ref{fig:box_qualitative}. Only the policy with full \name information (ours) can consistently perform the box packaging task with both non-perforated and perforated initial bands. For the first stage -- selecting the correct band for the packaging task -- all other policies adopt a singular strategy, either discarding or directly using the initially provided band every time, regardless of its suitability for packaging. Because the bands are difficult to visually distinguish, we hypothesize that each policy commits to an arbitrary trajectory mode from the data. However, the policy with full \name information can successfully select the appropriate band based on the tension perceived by the tactile skin.

Beyond allowing the policy to discern the suitability of a given elastic band, access to full \name information improves the robot's ability to dynamically wrap the band around the box. This can be seen in the fraction of rollouts that succeed in the band selection stage but fail in completing the packaging task by wrapping the band. While only $32\%$ and $70\%$ of rollouts for the \textit{No tactile} and \textit{\name (inner cols. only)} policies complete the second stage given success in the first stage, the policy with complete \name readings does so $86\%$ of the time.

\newcommand{\taskstageone}{Select}
\newcommand{\taskstagetwo}{Wrap}

\begin{table*}[t]
\centering
\small

\begin{tabular}{@{}l >
{\centering\arraybackslash}p{1.5cm} >{\centering\arraybackslash}p{1.5cm} >{\centering\arraybackslash}p{1.5cm} >{\centering\arraybackslash}p{1.5cm} >{\centering\arraybackslash}p{1.5cm} >{\centering\arraybackslash}p{1.5cm}@{}}
\toprule
\multirow{3}{*}{Model} & \multicolumn{2}{c}{\textbf{Pen reorientation}} & \multicolumn{2}{c}{\textbf{Box:} Non-Perforated Band} & \multicolumn{2}{c}{\textbf{Box:} Perforated Band} \\
\cmidrule(lr){2-3} \cmidrule(lr){4-5} \cmidrule(lr){6-7}
& No perturb & Perturb & \taskstageone & \taskstagetwo & \taskstageone & \taskstagetwo \\
\midrule
No tactile    & $\mathbf{19/20}$ & $0/20$ & $0/20$        & $0/20$        & $\mathbf{19/20}$ & $6/20$ \\
DIGIT~\cite{lambeta2020digit} &  $\mathbf{19/20}$ & $0/20 $ & $\mathbf{20/20}^*$        & $\mathbf{20/20}^*$        & $0/20$ & $0/20$ \\
DexSkin (ablated)    & $12/20$ & $0/20$ & $\mathbf{19/20}$ & $14/20$       & $1/20$         & $0/20$ \\
DexSkin (ours)                & $\mathbf{19/20}$ & $\mathbf{19/20}$ & $18/20$       & $\mathbf{17/20}$ & $\mathbf{19/20}$ & $\mathbf{15/20}$ \\
\bottomrule
\end{tabular}

\caption{
\label{tab:pen_box_packaging_results}
\small Pen reorientation and box packaging success rates. Policies with full \name information achieve the most consistent performance across tasks requiring high sensing coverage. For box packaging, we report success in two cumulative stages for evaluation only: (1) determining the correct elastic band to use and (2) physically wrapping the band around the box. 
$^{\boldsymbol{*}}$Placing the band around DIGIT's bulkier geometry elongates the band more, making wrapping it around the box easier. }
\vspace{-1em}
\end{table*}

\vspace{-0.5em}
\subsection{Calibration and Model Transfer Across Sensor Instances}
\label{sec:calibration_experiments}

The physical mechanisms and intricate fabrication procedures of tactile sensors make it challenging to manufacture hardware that produces identical sensing outputs given the same stimuli. For example, optical sensor readings vary based on the elastomer interface's texture, while outputs of both magnetic and resistive sensors rely on consistent filler loading and composite thickness~\cite{bhirangi2024anyskin, piezoresistive_sensitive}.
Yet, learned models that ingest tactile information are highly sensitive to the distribution of sensor signals present in the training data.
This presents a major practical bottleneck for using tactile sensing in robot learning. Because sensors must eventually be replaced, significant distribution shifts from existing training data often rendering existing models unusable. 

In this section, we present a calibration procedure for \name, and evaluate whether it can alleviate these challenges to enable downstream policy transfer to a different sensor hardware instance.

\textbf{Calibration procedure.} 
We develop two calibration protocols. The first houses \name with a custom 3D-printed airtight PLA chamber with a thin, soft Ecoflex 00-50 inner membrane, and ramps the relative internal pressure linearly from 0psi to 6psi, imposing uniform stress across the entire sensor surface. Fitting each taxel's response to an exponential curve yields both a forward mapping ($\Delta C/C_0 \to$  pressure) and an inverse mapping that when combined can align new sensor data with a legacy sensor’s output. The second involves mounting \name on a motorized linear stage with a force gauge and recording three loading-unloading cycles for each taxel to map sensor outputs to normal force. Details on calibration procedures and performance are in Appendix~\ref{appendix: sensor calibration}.

\begin{wraptable}{r}{0.42\textwidth}
\centering
\small
\vspace{-1em}
\begin{tabular}{lrr}
\toprule
Sensor configuration & Stage 1 & Stage 2 \\
\midrule
\multicolumn{3}{l}{\textbf{\name}} \\
Source sensors & $20/20$ & $20/20$ \\
Swapped (no calib.) &  $17/20$ & $12/20$ \\
Swapped (calib.) & $18/20$ & $16/20$ \\
Replaced (no calib.) &  $13/20$ & $5/20$ \\
Replaced (calib.) & $18/20$ & $14/20$ \\
\midrule
\multicolumn{3}{l}{\textbf{DIGIT}}\\
Source sensors  & $20/20$ & $0/20$\\
Swapped  & $0/20$ & $0/20$ \\
Swapped (diff. img)  & $20/20$ & $0/20$ \\
Replaced (diff. img) & $15/20$ & $0/20$ \\
\bottomrule
\end{tabular}
\caption{\small \label{tab:policy_transfer_experiments}Pen reorientation policy performance when transferred across sensor hardware. We report successes across two stages: (1) successfully reorienting the pen the first time and (2) detecting and fixing human perturbation.}
\vspace{-1.5em}
\end{wraptable}

\textbf{Policy transfer experiment.} We revisit the pen reorientation task with perturbation from Section~\ref{sec:pen_task}. We start with a policy trained on $50$ demonstrations from one pair of sensors, denoted \textit{Source sensors}, and test its performance when transferred to the same sensors but swapped between left and right fingers (\textit{Swapped}) and transferred to replacement sensors fabricated in a different batch (\textit{Replaced}). We test policy performance with and without calibrating sensor outputs.

The results are presented in Table~\ref{tab:policy_transfer_experiments}. \name exhibits reasonable policy transfer to target sensor configurations even without explicit calibration, and achieves an additional performance improvement after a short calibration procedure. 
In contrast, a policy trained on DIGIT sensors cannot transfer at all to a setting where gels have been swapped from their assignments during data collection, due to changes in visual appearance of the tactile readings. Performance improves if the policy ingests difference images relative to contactless readings, but still underperforms \name's calibration pipeline for replaced sensors.

\subsection{Real-World Online Robot Learning with \name}
\label{sec:rl_experiment}
Beyond applications in imitation learning, tactile sensing is a promising source of signals for robots that learn by interacting with their environments, for instance via reinforcement learning (RL). However, many tactile sensors lack the durability for repeated trial-and-error rollouts in the real world, and further, it is often unclear how to define reward functions based on sensor outputs from optical or magnetic mechanisms without collecting extensive datasets for supervision or calibration. 

In this section, we validate \name's suitability for online learning applications, by studying the task of delicate object grasping via real-world residual reinforcement learning.
\begin{wrapfigure}[20]{rt}{0.58\textwidth}
    \centering 
    \vspace{-1em}
    \includegraphics[width=\linewidth]{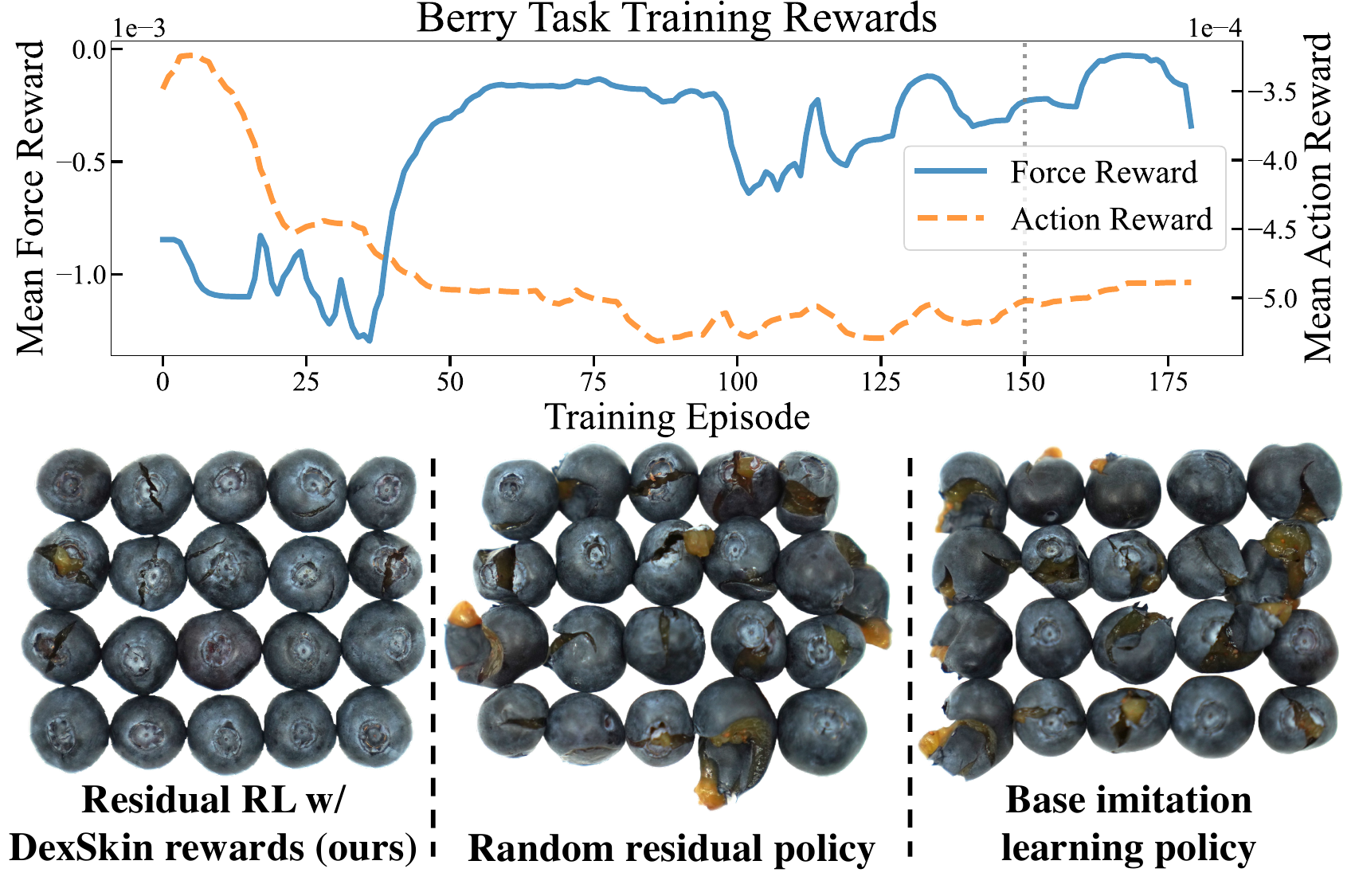}
    \caption{
    \small
        (Top) Residual policy learning curves. Over the course of training, the policy learns to minimize \name-output forces that exceed the specified threshold (maximizing \textit{force reward}), and trade off the action norm cost of larger residual actions (decreasing \textit{action reward}). We train with a faux berry until episode $150$, and real berries for the last $30$ episodes. 
        (Bottom) Berries after grasp and transport by each policy. The finetuned residual policy causes significantly less damage. 
    }
    \label{fig:berry_rl_results}
\end{wrapfigure}
Specifically, the robot's task is to grasp and transport a fragile blueberry without damaging it. This is challenging: episodes are over $200$ timesteps long, and a single overly forceful action can crush the delicate berry.

Given a policy trained via imitation on data from a gripper without a tactile sensor, we sensorize the gripper with \name, and then learn a neural network residual policy that parameterizes a modification of the frozen base policy's output.
The reward function penalizes tactile sensor readings above a set threshold, representing undesirably strong contacts, $\ell_2$ action magnitude, and failed grasps or object transports.
For real-world sample efficiency, we use soft actor-critic~\cite{haarnoja2018soft, stable-baselines3} to learn the residual policy. We do not advocate for a particular RL training procedure, but rather aim to study the capabilities of \name in such frameworks. 
Further experimental details are in Appendix~\ref{appendix:rl_details}.

\begin{wraptable}{r}{0.48\textwidth}
\setlength{\tabcolsep}{3pt} %
\renewcommand{\arraystretch}{0.95} %
\small %
\vspace{-1.5em}
\resizebox{\linewidth}{!}{\begin{tabular}{lrrr}
\toprule
& \makecell{Avg.\\Pressure\\(Artificial)} $\downarrow$ 
& \makecell{Avg.\\Pressure\\(Real)} $\downarrow$ 
& \makecell{Intact\\(Real)} $\uparrow$ \\
\midrule
Base IL (no tac.)  & 14.5kPa & 3.36kPa & 20\% \\
Random resid. & 6.17kPa &  3.64kPa  & 10\% \\
Resid. RL (ours) & \textbf{1.53kPa} & \textbf{1.92kPa} & \textbf{60\%} \\
\bottomrule
\end{tabular}}
\caption{\small \label{tab:berry_residual_results}  Evaluation of the final residual RL policy compared to base IL and uniformly random residual policy baselines. We report mean normal pressure applied per \name taxel across time over evaluation trajectories on the artificial berry used for training and real blueberries, and the percentage of \textbf{fully intact} real blueberries after transport for each policy. 
We perform $20$ evaluation trials per policy. 
} 
\vspace{-2em}
\end{wraptable}

Figure~\ref{fig:berry_rl_results} shows force and action rewards throughout the learning procedure, indicating that the residual policy learns to adjust the base policy's actions to produce gentler grasping behavior. 
Qualitatively, the base imitation learning policy (without tactile information) causes the robot to grasp the berries firmly and crush them -- likely as haptic feedback is absent for demo collection.
With the combined residual and base policy, the robot can grasp berries delicately enough to avoid breaking them. This is supported quantitatively by our experimental evaluation, shown in Table~\ref{tab:berry_residual_results}. 

This study illustrates how \name can not only be used to sensorize and adapt policies for agents trained initially without tactile information, but also that it enables sensible reward definitions without learned classifiers or calibration.

\vspace{-1em}
\section{Conclusion}%
\vspace{-1em}

In this paper, we introduced the \name framework, beginning with our novel elastic tactile sensor that conforms to different geometries, offering repeatable, localized, and high-coverage sensing. We equip a robotic gripper with \name and investigate its suitability for robotic learning applications. Through our experiments, we find that \name can enable imitation learning of improved policies for contact-rich manipulation tasks requiring high-coverage and localized sensing, enable learned model transfer across hardware instances via calibration, and provide feedback for real-world online reinforcement learning. We hope \name is a step toward tactile sensors that can rival human skin in coverage and sensitivity, while remaining practical and suitable for data-driven systems.

\section{Limitations} 

In this work, we instantiate \name on a pair of parallel jaw gripper fingers and conduct all experiments on this platform, which may not reflect its performance for different robot morphologies. 
Due to \name's customizability, sensorizing robots with morphologies like dexterous hands and covering even non-end-effector regions is an exciting and realizable next step. 

Additionally, during policy learning, we treat \name readings na\"ively as a one-dimensional feature vector. Future work may investigate how policy architectures can instead take advantage of spatial correlations in the \name data to learn more efficiently and robustly. 
On the sensor development, despite significant improvement on coverage, a blind spot of 66$^{\circ}$ measured in angular coverage remains. Furthermore, an electrical limitation exists, as ensuring a common ground is critical to maintaining signal integrity, particularly when the sensors are installed onto unshielded, non-metal grippers. The current open-sourced PCB design accommodates this with external jumper wires, which represents a less robust solution. To overcome this limitation, dedicated sensor shielding layers will be implemented in future iterations.

\clearpage
\acknowledgments{
We thank Neil Nie, Weiyu Liu, and Yunfan Jiang for their assistance with the robot and data collection setup, Annika Yong, Jonathan Peng, and Junyi Zhao for their assistance with the sensor characterization and calibration setup, and the members of the CogAI group for insightful discussions. This work is in part supported by NSF RI \#2338203, ONR MURI N00014-22-1-2740, ONR MURI N00014-24-1-2748, and the Stanford Robotics Center. ST was supported by NSF GRFP Grant No. DGE-1656518. BS, CX and ZB acknowledge support from Yi Man Liu and Sai Fu Fung, the Tianqiao and Chrissy Chen Ideation and Prototyping Lab, and Stanford Wearable Electronics Initiative (eWEAR) seed funding. 
}

\bibliography{cite}  %

\clearpage
\appendix

\textbf{In addition to the following information, please see the supplementary video and project website (\url{https://dex-skin.github.io/}) for a visual walkthrough of the framework hardware, a demonstration of \name's tactile sensing capabilities, and example videos of rollouts from each policy on each experiment.}

\section{Additional Details on \name Framework}
\label{sec:further_sensor_characterization}

\subsection{Additional Sensorized Morphologies}

Unlike other sensors limited by the form factors of commercial sensing elements, DexSkin allows for complete customization of the size, distribution, and layout of taxels. Its thin-sheet form also seamlessly integrates with existing designs.

\begin{figure}[h]
  \centering
  \vspace{.4em}
  \includegraphics[width=0.7\linewidth]{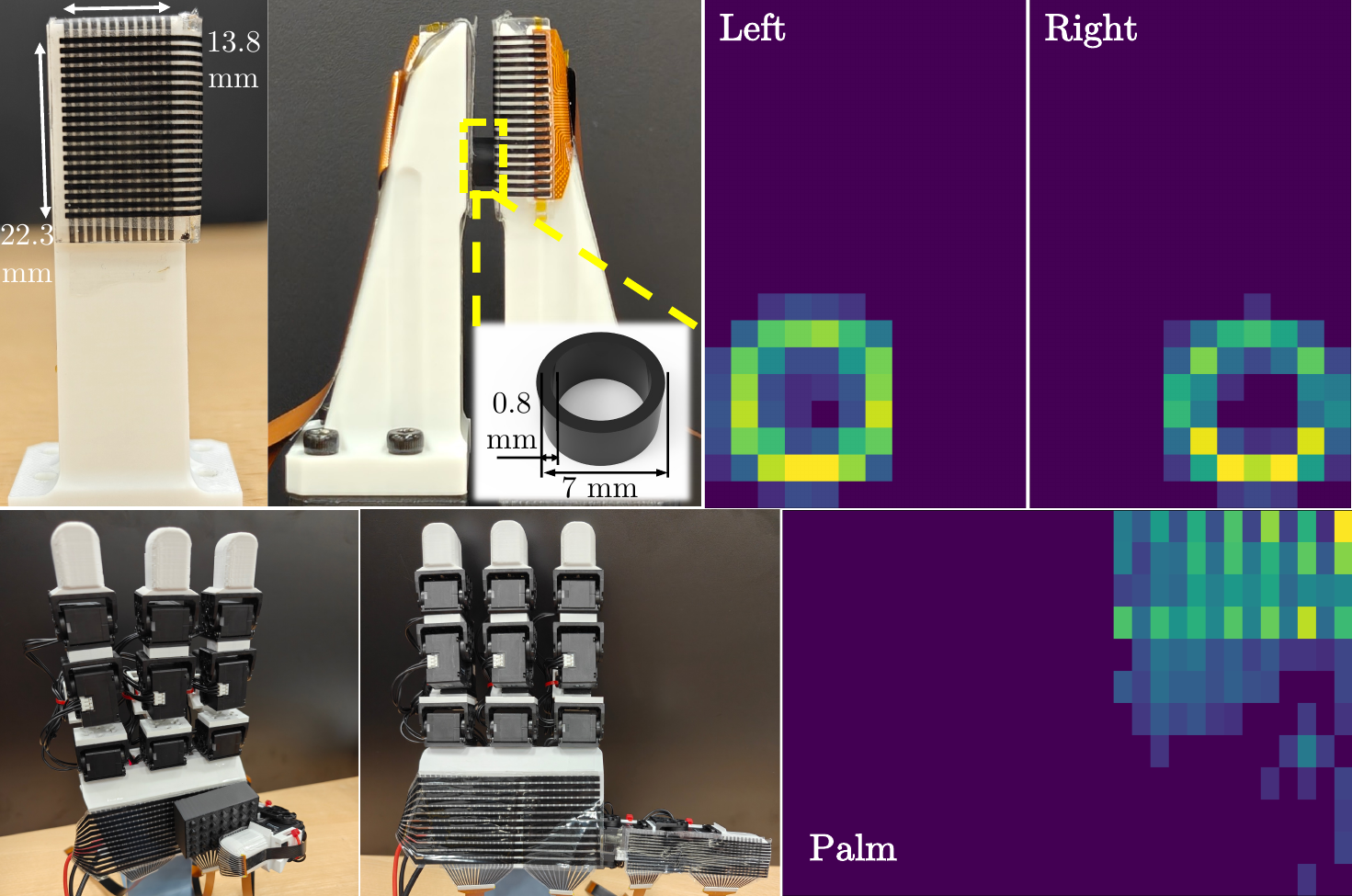}
  \caption{DexSkin application and example readings for 
   a (\textbf{top}) flat parallel jaw gripper (228 taxels each)  and (\textbf{bottom}) LEAP hand (372 taxels each for palm and thumb). For the LEAP hand, we visualize the palm taxels when holding the studded rectangular plastic block shown in the bottom left. 
  \label{fig:more_morphologies}}
\end{figure}

To showcase \name's tailorability, Figure~\ref{fig:more_morphologies} shows applications of DexSkin to 
additional morphologies: a LEAP robotic hand and a rigid parallel jaw gripper with planar grasping surfaces. For the LEAP hand, we sensorize the palm and thumb links. We omit the visualization for the thumb links here because the grasp configuration produced unclear contacts in that region. For the rigid gripper, we sensorize a $\SI{13.8}{\milli \meter}$ $\times$ $\SI{22.3}{\milli \meter}$ area on each fingertip.
    
\subsection{Sensor Characterization Setup}
\label{sec:vstage_setup}
We used a custom-built programmable pressure station to characterize the force-induced responses of our \name under repetitive loading and unloading cycles. The pressure station features a motorized vertical stage (Newmark Systems, NVS-12), a force gauge (Mark-10, M5-10), and either a precision LCR meter (Agilent E4980A) or our customized PCB for readout. \name is mounted on a 3D-printed rigid PLA core fixed to the stage with tape. 

Depending on the taxel to be tested, \name is either laid flat to orient one column of the cylindrical taxels normally upwards, or tilted to position a single dome taxel oriented normally upwards. The force gauge is fixed, while the stage moves in constant steps of \SI{0.01}{\milli \meter} to gradually increase and decrease the normal force applied as programmed. Cured 10:1 PDMS applicators (Dow Sylgard\texttrademark \space 184) matching the sizes of the tested taxels are placed between the force gauge probe and the surface of \name. They compensate for the rigidity and non-curving nature of the force probe, enabling conformal contacts with soft \name.

Measurements from the force gauge and the LCR meter are automatically synchronized with a custom LabView script at logging time. Measurements from the PCB are synchronized with the force gauge by aligning key timestamps.

\subsection{Sensor Characterization Experiments}
\label{sec:characterization_experiments}
To measure the physical characteristics of \name, we conducted key experiments to showcase its cross-taxel uniformity, low hysteresis, robustness, cyclic stability, broad sensing range, and minimal crosstalk. While we report the forces applied in the following sections, we would like to note that they are in terms of normal force per taxel and cannot be directly compared with other setups without accounting for differences in the size of force applicators. Therefore, in the following discussions, we normalize the forces to report normal pressure loads.

\subsubsection{\name's Cross-taxel Uniformity, Hysteresis and Cyclic Stability}
\label{sec:uniformity, hysteresis and cyclic stability}
\begin{figure*}[h]
    \centering
    \begin{subfigure}[b]{0.49\textwidth}  
        \includegraphics[width=\linewidth]{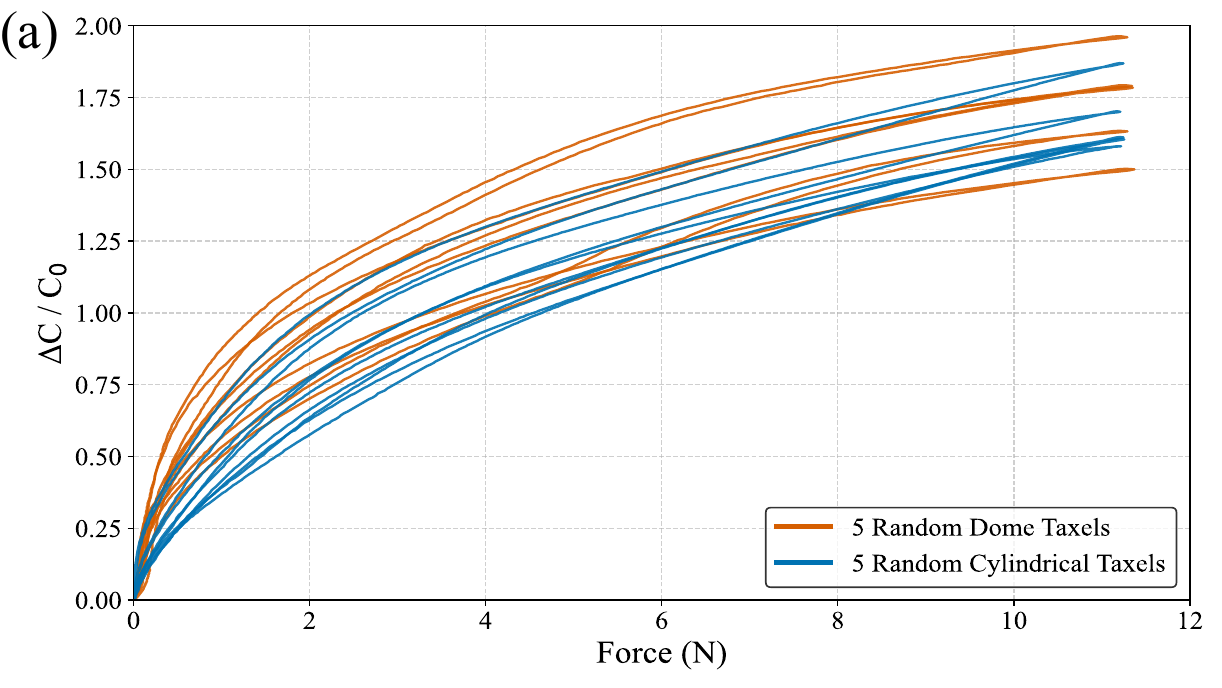}
        \label{fig:subfig1}
    \end{subfigure}
    \hfill
    \begin{subfigure}[b]{0.49\textwidth}  
        \centering\includegraphics[width=\linewidth]{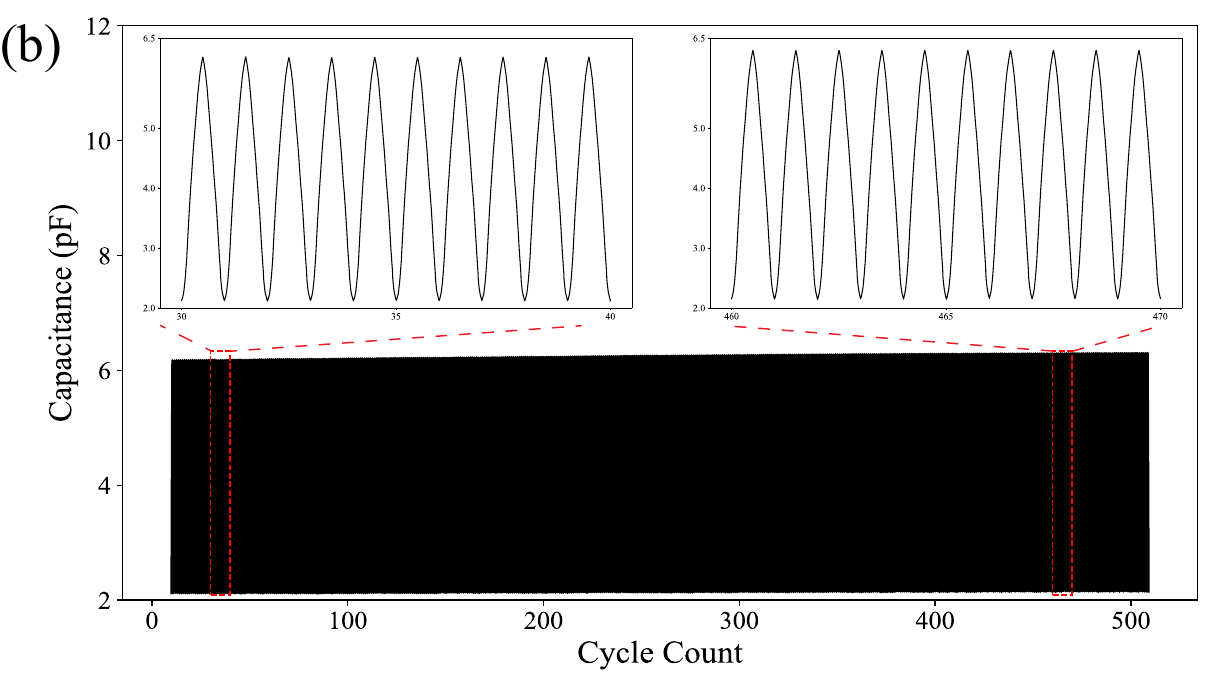}
        \label{fig:subfig2}
    \end{subfigure}
    \vspace{-1.5em}
    \caption{Characterization of the cross-taxel uniformity and cyclic stability of \name. (a) \textbf{Normalized capacitance change $\Delta C / C_{0}$ versus applied force for 10 randomly selected taxels} measured during loading and unloading up to $\SI{702.1}{\kilo\pascal}$ (\SI{11.12}{\newton}). Note the small hysteresis and high consistency across both dome and cylindrical taxels. (b) \textbf{Single-taxel cyclic loading up to \SI{702.1}{\kilo\pascal} (\SI{11.12}{\newton}) for 500 cycles.} The red rectangles (top) highlight 10 cycles (30th. to 40th., 460th. to 470th.) near the beginning and the end of the test, with zoomed-in view of consistent loading and unloading behavior for each highlighted segment. The black rectangle (bottom) depicts the taxel's changes in capacitance across all 500 cycles, with smooth and minimally-tilted top and bottom edges. This indicates minimal drift in both the maximum and the baseline responses.}

    \label{fig:sensor_characterization}
\end{figure*}

We evaluated the sensor's cross-taxel uniformity by randomly selecting ten taxels (five on the dome and five on the cylindrical body) and measuring their responses subject to normal pressure up to  $\SI{702.1}{\kilo\pascal}$ (\SI{11.12}{\newton}), as visualized in Figure~\ref{fig:sensor_characterization}(a). All ten taxels, though designed with different sizes, exhibit consistent logarithmic force-capacitance (normalized) responses in the entire sensing range. Additionally, the hysteresis performance, indicated by the maximum height of the gap between the loading and unloading curves, averages 6.52\% $\pm$ 1.58\% across ten selected taxels, is significantly lower than the 17.8\% reported for the STAG glove \cite{sundaram2019STAG} and comparable to commercial flexible force sensors (Tekscan FlexiForce WB201).

We then evaluated the long-term cyclic stability and robustness of \name by subjecting a single taxel from the sensor dome to 500 cyclic loading and unloading cycles of up to $\SI{702.1}{\kilo\pascal}$ (\SI{11.12}{\newton}) normal loading, over the course of eight hours. As depicted in Figure~\ref{fig:sensor_characterization}(b), the smooth and minimally tilted response over 500 cycles shown at the bottom indicates minimal drift over time, and quantitative analysis reveals a peak drift of 2.09\% and a zero drift of 1.72\%. Zoomed-in views of the beginning and ending cycles further confirm consistent loading and unloading behaviors.
This level of consistency and stability of \name, attainable without the need for individual calibration or drift compensation, enables us to directly input sensor responses into our policy learning pipeline. 

\subsubsection{\name's Broad Sensing Range}
\label{sec:broad_sensing_range}
To support diverse use cases, we evaluated our \name under two sleeve configurations: a soft low-infill TPE sleeve for compliant, low-load tasks and a more rigid sleeve with 100\% TPE infill for high-load applications. We conducted cyclic loading tests using our custom-designed PCB for readout, in the same configuration as deployed on the real robotic system, to characterize each configuration's response to normal pressure loads. We used the vertical-stage based sensor characterization setup discussed in Section~\ref{sec:vstage_setup} to apply cyclic normal pressure loadings on a randomly selected taxel. 

\begin{figure}[t]
    \centering \includegraphics[width=1.0\columnwidth]{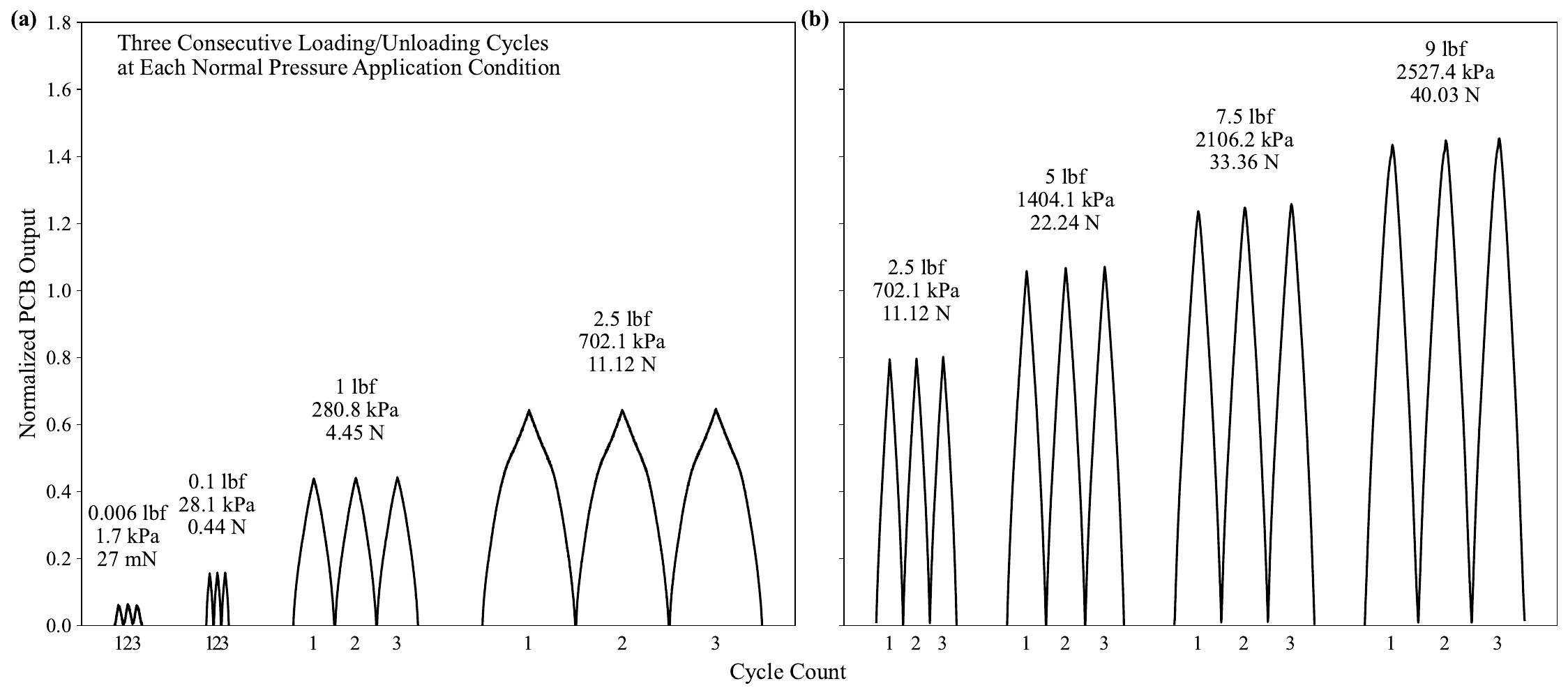}
    \caption{
    Normalized PCB output from a randomly selected taxel during three consecutive loading and unloading cycles under different normal pressure application conditions for two sleeve types.
    (a) \name with the low-fill TPE soft sleeve: data shown for $\SI{1.7}{\kilo \pascal}$ (\SI{27}{\milli \newton}), $\SI{28.1}{\kilo \pascal}$ (\SI{0.44}{\newton}), $\SI{280.8}{\kilo \pascal}$ (\SI{4.45}{\newton}), and $\SI{702.1}{\kilo\pascal}$ (\SI{11.12}{\newton}). (b) \name with a more rigid 100\% infill TPE sleeve: data shown for $\SI{702.1}{\kilo\pascal}$ (\SI{11.12}{\newton}), $\SI{1404.1}{\kilo\pascal}$ (\SI{22.24}{\newton}), $\SI{2106.2}{\kilo\pascal}$ (\SI{33.36}{\newton}), and $\SI{2527.4}{\kilo \pascal}$ (\SI{40.03}{\newton}). Taxel readings are normalized with respect to the initial zero-load values. The y-axis, representing timestamps, has been relabeled as cycle counts to better highlight the three consecutive cycles performed in each condition. The vertical stage step size defaults to $\SI{0.01}{\milli \meter}$ but is reduced to $\SI{0.002}{\milli \meter}$ for the $\SI{1.7}{\kilo \pascal}$ (\SI{27}{\milli \newton}) condition to prevent force overshoot.
    }
    \label{fig:consecutive_loading}
\end{figure}

As shown in Figure~\ref{fig:consecutive_loading}, both configurations demonstrate stable and repeatable responses across three loading-unloading cycles, aligning well with the stability observed using an LCR meter. The soft sleeve variant of \name responds reliably to normal pressure loads as small as $\SI{1.7}{\kilo \pascal}$ (\SI{27}{\milli \newton}) and up to $\SI{702.1}{\kilo\pascal}$ (\SI{11.12}{\newton}), beyond which the sleeve structure yields to provide cushioning. The loading range of soft sleeve \name, characterized up to $\SI{702.1}{\kilo\pascal}$, exceed prior sensors such as Gelsight ($\sim\SI{200}{\kilo\pascal}$) \cite{yuan2017gelsighta} and DIGIT Pinki ($\sim\SI{80}{\kilo\pascal}$) \cite{di2024using}. The variant of \name with a rigid sleeve, in contrast, withstands higher loads without significant deformations and stably responds to normal pressure loads beyond those achieved by the soft sleeve counterpart, up to $\SI{2527.4}{\kilo \pascal}$ (\SI{40.03}{\newton}).

\begin{figure}[h]
    \centering \includegraphics[width=1.0\linewidth]{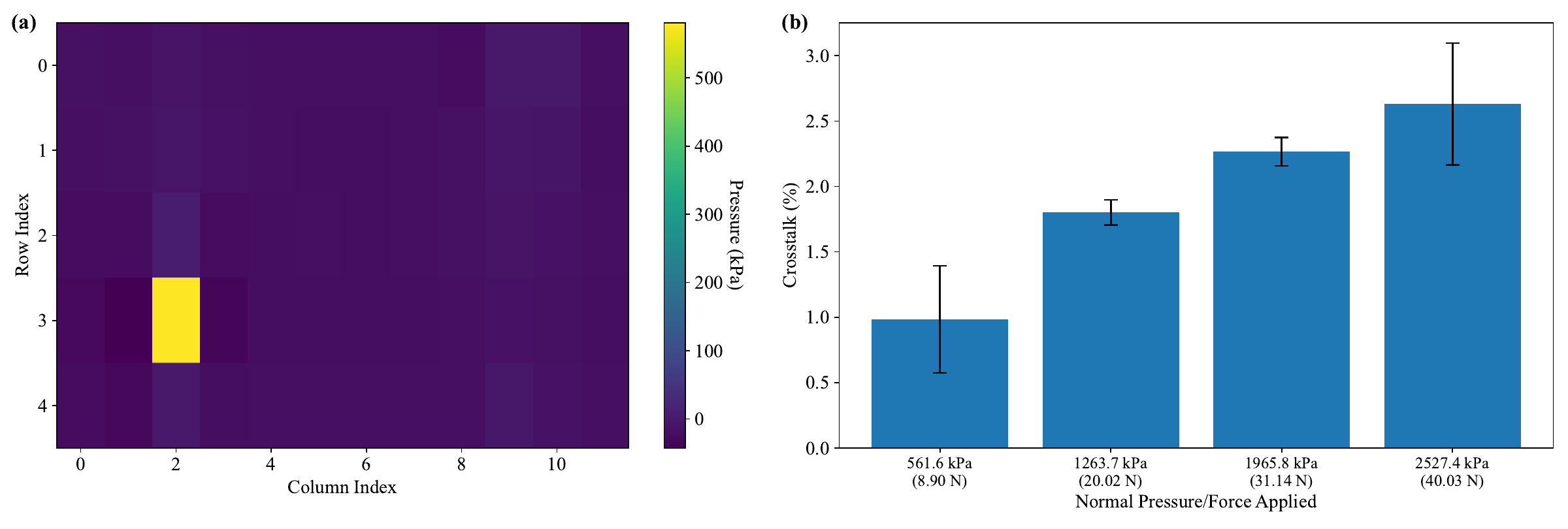}
    \caption{
    \label{fig:crosstalk_evaluation} Evaluation of \name's crosstalk under a localized normal load. (a) Heatmap of the sensed normal pressure across \name's full 60 taxels when a single taxel is subjected to a localized normal load measuring $\SI{561.6}{\kilo\pascal}$ (\SI{8.90}{\newton}). (b) Quantitative evaluation of the maximum crosstalk (in percentage) observed across the remaining 59 taxels with four different normal pressures applied on the loaded taxel. The average crosstalk remains below 3\% at the four sampled loads, demonstrating minimal interference and couplings between closely-positioned taxels.
    }
\end{figure}
\subsubsection{\name's Minimal Crosstalk}

We evaluated the level of crosstalk using the same vertical-stage based sensor characterization setup as described in Section~\ref{sec:vstage_setup} to apply normal loads ranging from $\SI{280.8}{\kilo\pascal}$ (\SI{4.45}{\newton}) to $\SI{2808.1}{\kilo \pascal}$ (\SI{44.48}{\newton}) on a randomly selected taxel in 100\% infill TPE sleeve \name. We read out the sensor outputs using our custom PCB in normal pressure applied. We were able to obtain 1435 force-sensor value pairs for crosstalk evaluations. 

Figure~\ref{fig:crosstalk_evaluation}(a) shows a representative heatmap of the sensed normal pressure when a single taxel encounters a normal load of $\SI{561.6}{\kilo\pascal}$ (\SI{8.90}{\newton}). The visualized response is highly localized and exhibits minimal activations in the surrounding taxels.

To quantify crosstalk, we define the  crosstalk (in \%) as the ratio between the highest pressure sensed by any of the 59 unloaded taxels and the pressure sensed by the loaded taxel: \[
\text{Crosstalk (\%)} = \left( \frac{\max\limits_{j \ne i} P_j}{P_i} \right) \times 100\%
\]
where $P_i$ denotes the pressure sensed by the loaded taxel. This formulation of crosstalk reflects the magnitude of ghost contacts at regions experiencing zero normal loads. As seen in Figure~\ref{fig:crosstalk_evaluation}(b), the average crosstalk increases with the magnitude of normal loads but remains below 3\% for all four loading conditions shown. Expanding the analysis to all 1435 pairs acquired, we obtain a mean crosstalk of $1.48\% \pm 1.07\%$. This highlights the minimal amount of interference and couplings and the individually addressable characteristic of taxels in \name, which is required for accurately localizing contact events, particularly for densely-packed taxel arrays.

\subsection{Readout circuitry}
\label{sec:readout circuitry}
\begin{figure}
    \centering
    \includegraphics[width=0.5\linewidth]{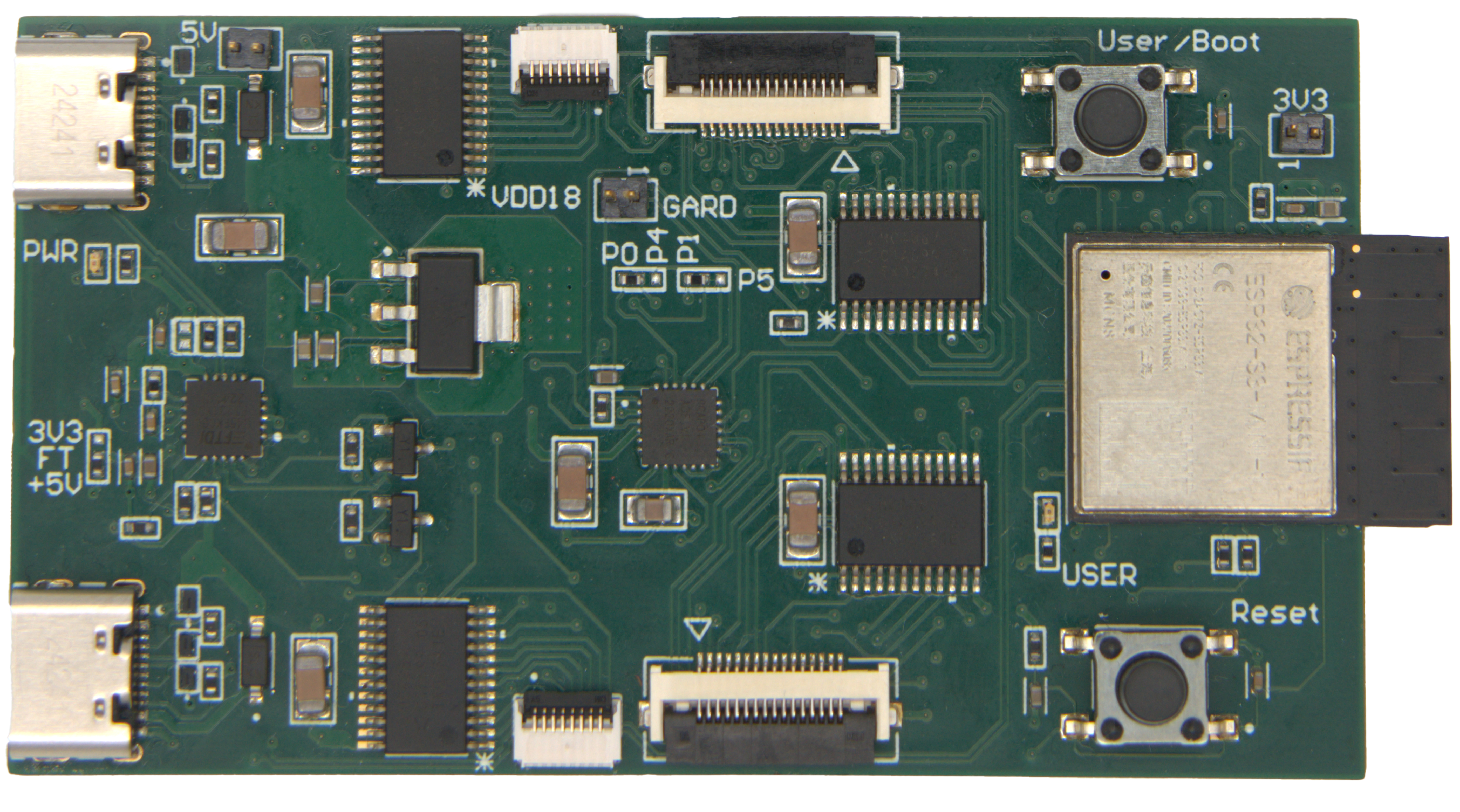}
    \caption{Custom readout PCB used by \name.}
    \label{fig:pcb}
\end{figure}

We measure pressure-induced capacitance changes with an inexpensive (costs USD \$18.4 at quantities of 1000) custom readout PCB shown in Figure~\ref{fig:pcb}, measuring \SI{72.6}{\milli\meter} x \SI{40.6}{\milli\meter}. A single PCB is capable of handling inputs from all 120 taxels across both fingers, sufficient for a parallel jaw gripper setup. The onboard ESP32-S3 microcontroller scans through each taxel sequentially using multiplexers while communicating with the onboard capacitance-to-digital chip to clock discharging for capacitance measurement. The PCB also provides active and passive shielding options to minimize the effects of electromagnetic interference (EMI). During our experiments, each taxel is measured four times and averaged for accuracy. The resulting serial data stream is transmitted to the computer at \SI{30}{\hertz}.

\subsection{Sensor Calibration with Readout PCB}
\label{appendix: sensor calibration}
\name exhibits consistency and low hysteresis, which are both crucial for reliable calibrations that accurately map measured sensor outputs into normal force and pressure values. In the following sections, we describe a calibration method that builds an accurate mapping from the PCB outputs to the normal pressure and force experienced. We then showcase the method by calibrating five randomly selected taxels and evaluating their force estimation error post-calibration.
\subsubsection{Comprehensive Normal Force Calibration}
\label{sec:force_calibration_method}
For calibrating \name's output to normal forces, we employ the same vertical-stage-based sensor characterization setup introduced in Section~\ref{sec:vstage_setup} with max normal load set to $\SI{157.8}{\kilo\pascal}$ (\SI{2.5}{\newton}), step size set to $\SI{0.01}{\milli \meter}$, and number of loading-unloading cycles set to 3. \name taxels consistently follow an exponential pattern between the PCB readout and the force applied; thus we select the exponential function $a \cdot \left( e^{b \cdot (x+d)} - e^{b \cdot d} \right)$ where a, b, and d are the three fitting parameters to describe the relationship. After four loading/unloading cycles, we obtain the dataset required for our parameter fitting process. We first detect the peaks of these cycles and align the peak timestamps between the PCB readings and the applied force before performing downsampling on the former to match the sampling frequency of the force gauge. The resulting downsampled dataset is then used to fit the exponential trend line and derive the three fitting parameters. This completes the calibration of one taxel and takes approximately 3 minutes. The entire calibration process for all 60 taxels on \name takes around 3 hours.

\subsubsection{Force Estimation Performance of Calibrated-Taxels}
\begin{figure}[t]
    \centering \includegraphics[width=0.5\columnwidth]{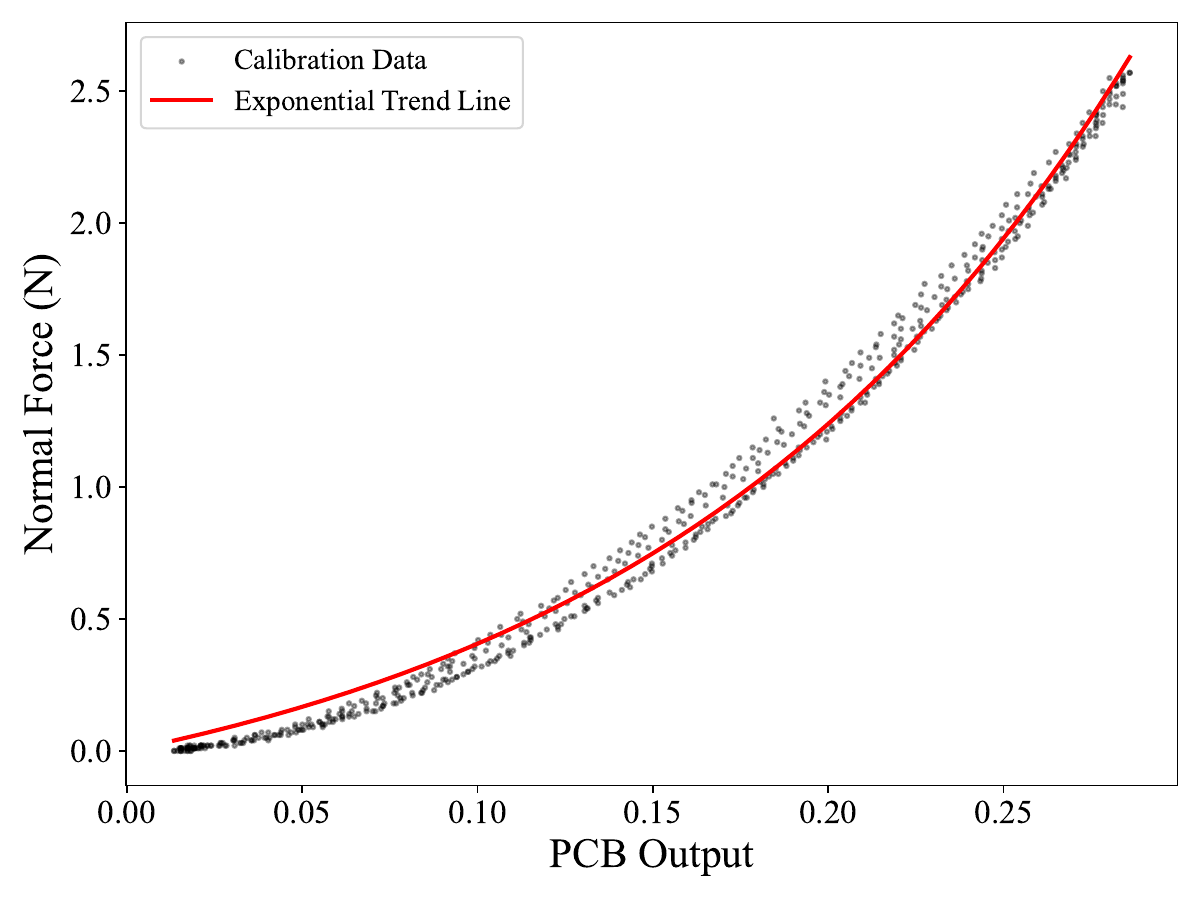}
    \caption{
    Representative scatter plot visualizing the collected force estimation dataset with 4 loading-unloading cycles with normal force applied up to \SI{2.5}{\newton} ($\SI{157.8}{\kilo\pascal}$) on a randomly selected taxel. Each of the gray dots represents a force-sensor output pair obtained during the calibration process, and the red line depicted is the exponential trend line obtained by fitting onto the calibration dataset composed of gray dots. The $R^2$ value for the trend line is 0.9929, indicating an excellent fit.
    }
    \label{fig:exponential trend line}
\end{figure}

\begin{figure}[t]
    \centering \includegraphics[width=0.5\columnwidth]{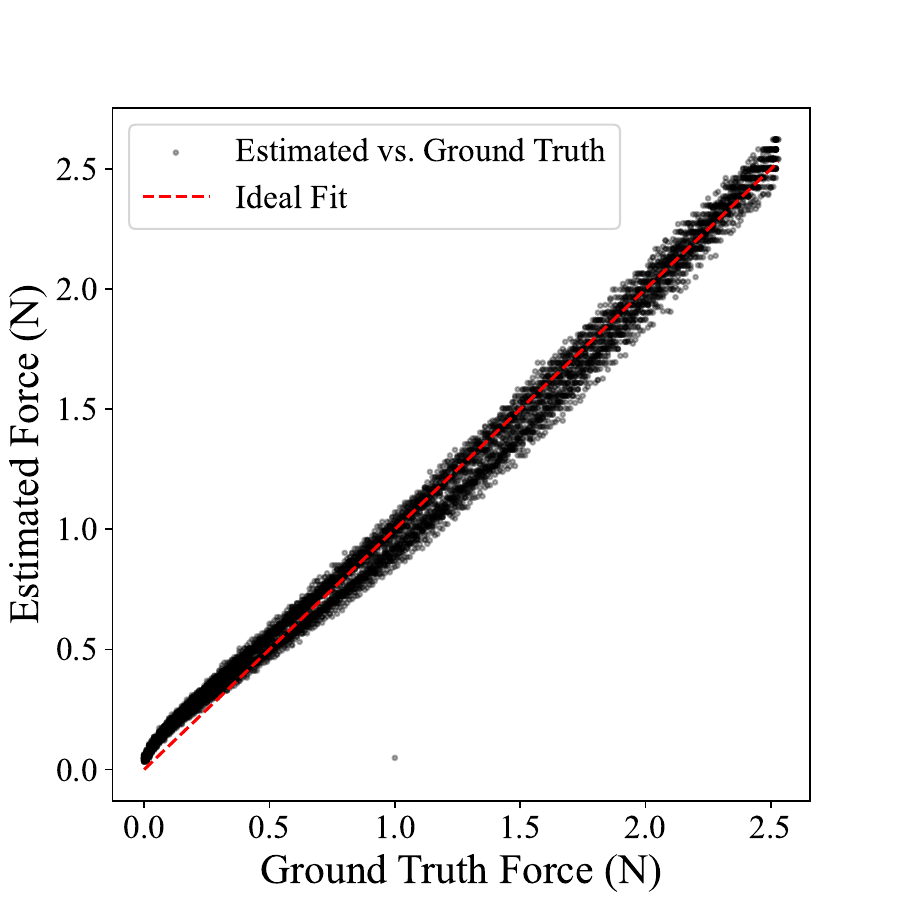}
    \caption{
    Representative scatter plot of estimated vs. ground truth force for one of the randomly selected taxels. We visualize the force estimation results against 7367 collected ground truth normal forces up to \SI{2.5}{\newton} ($\SI{157.8}{\kilo\pascal}$). Each of the gray dots represents an estimated force-ground truth force pairs and the red line depicted is the ideal fit line representing perfect estimation.
    }
    \label{fig:force_prediction}
\end{figure}

To evaluate force estimation errors, we randomly selected five taxels, calibrated them following procedures outlined in the previous section, and obtained the three fitting parameters for its exponential trend lines. The exponential calibration curve for one of the taxels is visualized in Figure~\ref{fig:exponential trend line}. Then we used the vertical stage setup to apply normal loads up to \SI{2.5}{\newton} ($\SI{157.8}{\kilo\pascal}$) onto each of the five taxels until we have obtained more than 6500 pairs of ground truth normal force values and PCB readings for each of them. For each resulting evaluation dataset, we feed the PCB readings into the fitted exponential trend line for force estimation, the results of which for one of the five taxels are plotted with the ground truth force values in Figure~\ref{fig:force_prediction}. Quantitatively, we collected a total of 39304 force-PCB value pairs across taxels and report \SI[separate-uncertainty]{0.086 \pm 0.021}{\newton} as the root mean square error for estimated forces. 

\subsubsection{Three-Minute Pneumatic Pressure Calibration for Policy Transfer}

In real-world robotic deployments, tactile sensors inevitably suffer from physical damage and wear, eventually failing. Yet due to fabrication variations, replacing a failed unit often results in distinct force responses which invalidates previously collected data and trained neural networks. 

\begin{figure}[t]
    \centering \includegraphics[width=1.0\columnwidth]{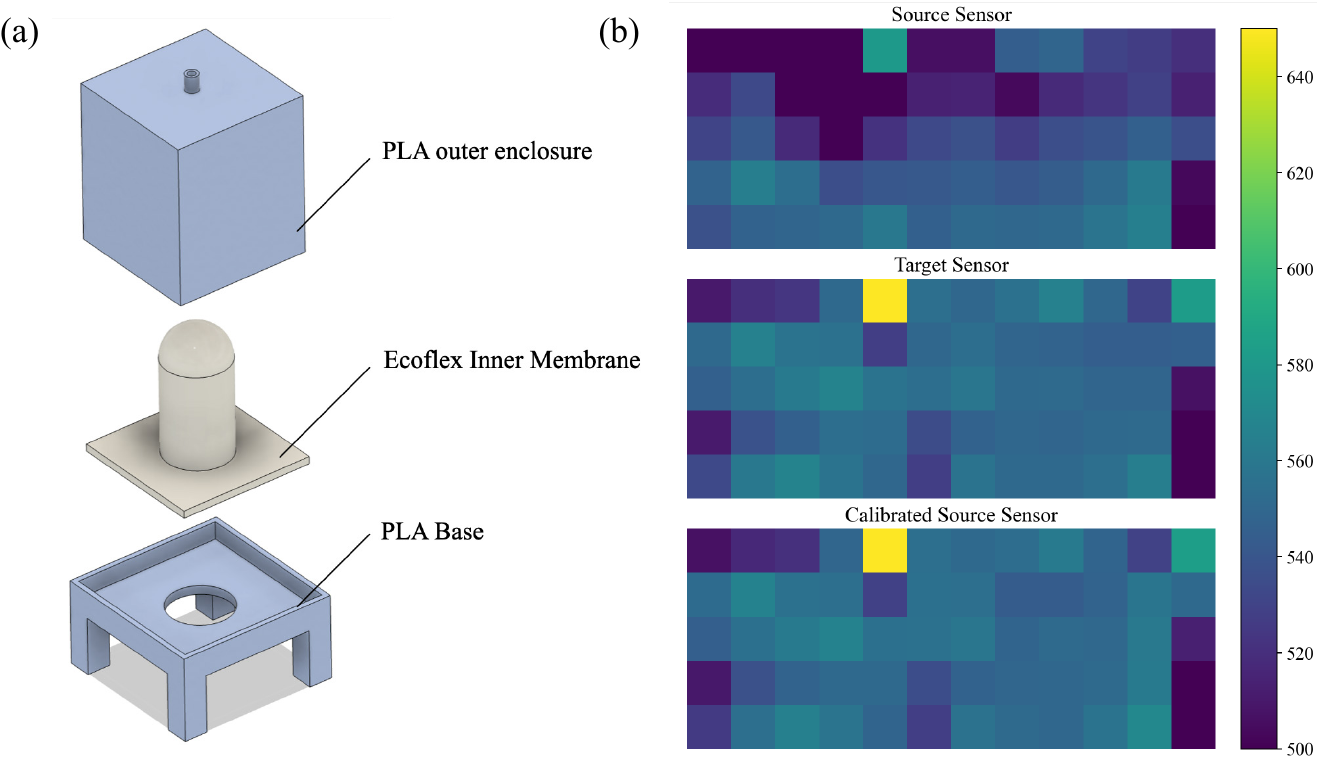}
    \caption{
    The pneumatic normal pressure calibration setup for \name. (a) Exploded view of the airtight chamber that houses \name during calibration and applies stress over the entire sensing region when pressurized. (b) Example heatmaps showing a set of raw values from the source sensor, raw PCB values from the target sensor to be transferred to, and the source sensor values after pneumatic pressure calibration when subject to the same 18.7 kPa normal pressure load. Ideally, the target sensor and calibrated source sensor readings should match.
    }
    \label{fig:pneumatic_calibration}
\end{figure}

To address this, we developed a pneumatic calibration fixture, shown in Figure~\ref{fig:pneumatic_calibration}(a). It is comprised of a custom 3D-printed PLA chamber with an air outlet on top, a thin and soft Ecoflex 00-50 inner membrane adhered to the bottom surface of the chamber, and a 3D-printed PLA base with leg support tailored for \name's height. We adopt a breadboard setup where the onboard ESP32-C3 microcontroller drives a 2N7000 MOSFET in low-side configuration to modulate a 6V DC air pump for inflation and reads the chamber pressure from a parallel-connected gas pressure sensor (Honeywell ABP) for feedback. When pressurized, the internal membrane deforms and applies a uniform stress over the surface of the \name sensor. The system is capable of ramping up the relative chamber pressure up to $6psi$ ($\SI{41.4}{\kilo\pascal}$).

Similarly to Section~\ref{sec:force_calibration_method}, with the pressure and corresponding sensor values collected, one can fit an exponential function $a \cdot e^{b \cdot x} + d$ where a, b, d are the three fitting parameters. For the pen reorientation model transfer evaluation, we capped the maximum inflating pressure at $\SI{18.7}{\kilo\pascal}$.

With the three fitting parameters obtained for both the source sensor and the target sensor, we adopt the following formula to calculate the estimated target sensor outputs using outputs from the source sensor:
\[
C_{2}
= \frac{C_{0,2}}{b_{2}}
  \ln\!\Biggl(
    \frac{
      a_{1}\,\exp\!\Bigl(b_{1}\,\frac{C_{1} - C_{0,1}}{C_{0,1}}\Bigr)
      +\,d_{1} - d_{2}
    }{a_{2}}
  \Biggr)
  +\,C_{0,2}
\]
where $a_{1}, b_{1}, d_{1}$ are fitting parameters for the source sensor, $C_{1}, C_{0,1}$ are the current output and no-load output for the source sensor respectively, $a_{2}, b_{2}, d_{2}$ are fitting parameters for the target sensor, and $C_{2}, C_{0,2}$ are the current output and no-load output for the target sensor respectively. 

To evaluate the effectiveness of the pressure calibration, we subject two \name fingers to $\SI{18.7}{\kilo\pascal}$ pressure loads and obtained the fitting parameters. Then we converted the recorded source sensor outputs using the formula shown above. Under a relatively uniform load of $\SI{18.7}{\kilo\pascal}$, the outputs for the source sensor, the target sensor, and the source sensor calibrated to emulate the target sensor are visualized in Figure~\ref{fig:pneumatic_calibration}(b). The heatmaps of the target sensor and the calibrated source sensor are nearly indistinguishable while the source sensor's map remains noticeably different. Quantitatively, the calibration brings the source sensor's outputs much closer to the target sensor's: the Structural Similarity Index Measure (SSIM) increases from 0.3344 to 0.9442, and the mean squared error decreases from 1085.1 to 13.9. The effectiveness of this calibration procedure can be seen not only in these direct comparisons of sensor output values but also by downstream improvements in policy model transfer performance (Section~\ref{sec:calibration_experiments}). 

\section{Additional Details on Robotic Experiments}
\begin{figure}[h]
    \centering \includegraphics[width=0.6\linewidth]{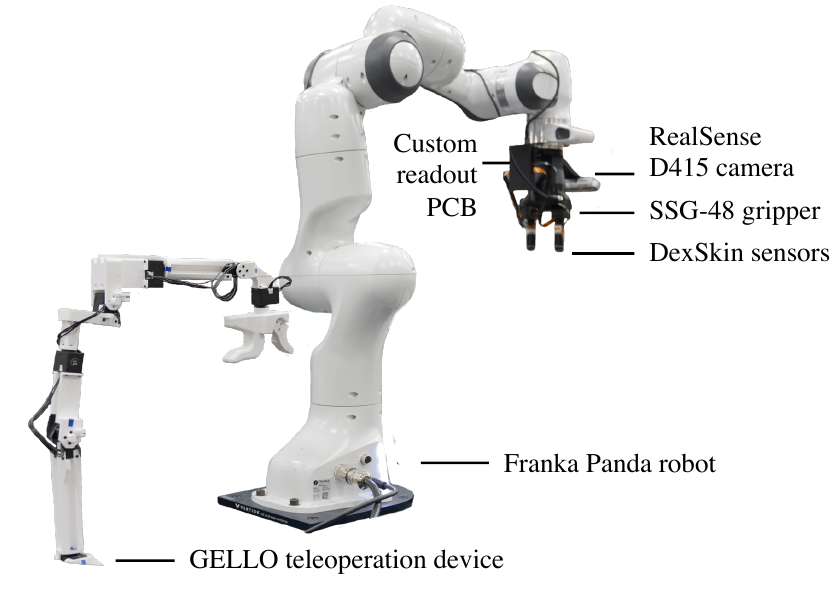}
    \caption{
    \label{fig:robot_setup} Hardware setup for manipulation and teleoperated demonstration collection with \name.
    Our system consists of a Franka Panda robot, GELLO teleoperation device~\cite{wu2024gello}, SSG-48 gripper with \name-equipped fingers, RealSense D415 RGB-D camera, and custom \name readout circuitry.
    }
\end{figure}
Figure~\ref{fig:robot_setup} shows an overview of the entire robotic hardware system, including the sensorized gripper, wrist camera, robot, and teleoperation device.

\subsection{Details on Imitation Learning for Manipulation with Expanded Coverage and Tailorability}
\subsubsection{Task Definitions}
\label{appendix:task_setup}
Here we provide additional details about the physical definitions for each task:

\begin{itemize}[left=1.5em]
    \item Pen reorientation: We use the same standard-size Expo pen when collecting all training episodes and when performing evaluation. The starting position of the pen on the table is essentially fixed (with allowances for small 1-\SI{2}{\centi \meter} variations) at the beginning of each episode. We also define a fixed target final location for the pen on the surface of the table. 
    \item Box packaging: We mount a plastic sushi box in a fixed location on the table. We also 3D-print a small holder to hold the extra elastic band. All bands that we use are of an identical \#19 size. 
\end{itemize}

\subsubsection{Policy Training and Inference}
To train imitation learning policies for the pen and box tasks in Section~\ref{sec:pen_task} and ~\ref{sec:calibration_experiments}, as well as the base berry picking policy in Section~\ref{sec:rl_experiment}, we use diffusion policies~\cite{chi2023diffusionpolicy, chi2024diffusionpolicy}. 
We use the open-source implementation by the original authors and default hyperparameters from the official release, specifically the \texttt{unet\_hybrid\_image} configuration specified in the GitHub repository. We summarize these hyperparameters in Table~\ref{tab:hyperparams}.

\begin{table}[h]
    \centering
    \begin{tabular}{lr}
        \toprule
        \textbf{Hyperparameter} & \textbf{Value} \\
        \midrule
        \multicolumn{2}{l}{\textbf{Model Configuration}} \\
        Horizon ($H$) & 16 \\
        Observation Steps ($n_{\text{obs}}$) & 2 \\
        Inference Steps ($n_{\text{action}}$) & 8 \\
        Crop Shape & [202, 202] \\
        Kernel Size & 5 \\
        \midrule
        \multicolumn{2}{l}{\textbf{Optimizer (AdamW)}} \\
        Learning Rate ($\eta$) & 1e-4 \\
        Betas ($\beta_1, \beta_2$) & (0.95, 0.999) \\
        Epsilon ($\epsilon$) & 1e-8 \\
        Weight Decay & 1e-6 \\
        \bottomrule
    \end{tabular}
    \vspace{1em}
    \caption{Diffusion policy training and inference hyperparameters.}
    \label{tab:hyperparams}
\end{table}

For policy configurations where \name observations are provided as input, the \name observations are processed as low-dimensional observations of dimension $120$, which are ingested by the policy in the same way as the robot proprioceptive state and gripper state, that is, using fully-connected (linear) layers. We perform random crops on the RGB camera observations, which is a standard recommendation for training image-based diffusion policies.

Training is carried out on a mixture of NVIDIA TITAN RTX, 3090, and A5000 GPUs. Training policies for the pen reorientation task takes approximately 12 hours, completing $581$k global steps. The policies for the box packaging task trained for approximately 34 hours, completing $554$k global steps.

At inference time, we perform $8$ diffusion policy denoising steps for computational efficiency, as recommended by \citet{chi2023diffusionpolicy}. We execute all $16$ actions output by the policy before performing another policy forward pass, rolling out the policy actions at 20Hz. 

We perform all policy inference on an NVIDIA RTX A4000 GPU.

\subsubsection{Evaluation Details}
When conducting comparative evaluations for a set of policies, we programmatically randomly shuffle the order in which the trials for each policy will be evaluated. This order is unknown to the experimental evaluator to eliminate potential bias. We perform experimental trials for all policies for a particular experimental configuration (e.g. no perturbation vs. with perturbation for the pen reorientation task) consecutively before moving to the next configuration. 

We define success conditions for each task as follows:
\begin{itemize}[left=1.5em]
    \item Pen reorientation: A trajectory is considered successful if the pen reaches an orientation that is within $30^\circ$ of perpendicular to the ground, and then hovers or rests its cap over a predefined target location. 
    In the experimental setting where the human experimenter perturbs the pen, the trajectory is deemed successful only if the robot reorients the pen to within $30^\circ$ of perpendicular to the ground initially and returns it to this pose after the human perturbs the pen one time to its starting orientation.

    In the calibration experiments described in Section~\ref{sec:calibration_experiments}, we report more granular notions of success: solving stage 1 involves completing a successful initial reorientation of the pen (prior to perturbation), and stage 2 requires detecting and fixing the human perturbation. 

    \item Box packaging: A trajectory is considered successful if both of the following conditions are true: (a) the robot discards the band it initially is holding if it is perforated and does not discard it otherwise, and (b) the trajectory ends with the rubber band wrapped around the box and lid, with the rubber band not touching any part of the robot. We also report performance on the ``Select Band'' subtask, which corresponds to satisfying condition (a).
\end{itemize}

\subsection{Details on Real-World Online Learning Experiments}
\label{appendix:rl_details}
\subsubsection{Reinforcement Learning Training Details}
To train the base imitation learning policy (``base policy''), we collect $50$ task demonstrations \textbf{on a robot equipped with an unsensorized gripper}. Using this data, we train a diffusion policy following 
 almost an identical configuration as the box packaging task in Section~\ref{sec:pen_task}, including the use of wrist camera inputs. However, the base imitation policy training has one key difference: while the experiments in Section~\ref{sec:pen_task} used \name tactile sensor readings as inputs, our base policy for online learning \textbf{does not} use tactile information as input.

Then, we use SAC~\cite{haarnoja2018soft, stable-baselines3} to train a residual policy. The residual policy takes as input the same proprioceptive information as the base policy, but \textbf{does not} use visual (wrist camera image) inputs and instead \textbf{does} use $120$ taxels of \name tactile information. Excluding visual inputs from the residual policy input keeps the residual policy lightweight and quicker to learn, and avoids potential visual distribution shifts from faux to real berries. In addition to the proprioceptive states from the robot, the residual policy receives the base policy's predicted action $a_b$ at a given step as input, such that it has access to the action it is modifying. 

Just as in the imitation learning experiments, the base policy computes a sequence of $16$ actions, which are all executed (with residual actions applied) before the base policy is queried again.

The residual policy's actions $a_r \in \mathcal{A_R} = [0.8, 1.2]$ (where $\mathcal{A_R}$ is the residual policy's action space) represent scaling modifications to the base policy's gripper action. That is, the action executed on the robot is modified for the gripper joint to instead be $a = \min (\max(a_b * a_r, 0), 1)$. This scaling formulation allows the residual policy to have a high level of control when the gripper is mostly closed, while reducing high-amplitude oscillations when the gripper is mostly open (e.g. when the robot is simply approaching the object) that cause exploration to become very challenging. 

We apply exponential moving average smoothing to the final gripper actions with $\alpha = 0.3$ to reduce high frequency jittering that makes exploration challenging and can potentially damage hardware.

During training, we train for $130$ episodes, approximately equivalent to $42$k environment steps. An episode ends after the robot attempts or successfully deposits a berry into the basket, returning its end effector to above a prespecified height, or reaches a time limit of $425$ steps (approximately 30 seconds) without doing so. For the first $100$ episodes, we train using a faux berry that has a visual appearance and geometry similar to that of a real berry. For the last $30$ episodes, to fine-tune the model for better performance on real blueberries, we remove all transitions from the replay buffer and then resume training using real blueberries.

The reward function has three terms as follows:  
\begin{itemize}[left=1.5em]
    \item \textbf{Large force penalty}: Unlike some optical or magnetic tactile sensors, \name readings are immediately localized and readily interpretable, and can thus be used to compute rewards such as this large force penalty, which equals the amount by which each \name taxel reading exceeds a specified threshold. We first filter out very large tactile values ($> 0.35$) that are caused by unintentional contacts with the surface of the plate that the berry starts on. That is, using $t \in \mathbb{R}^{120}$ to represent the \name outputs, 
    $$r_{force} = \| \max(0, t - t_{thresh})\|^2_2 $$
    where $t_{thresh} = 0.1$.
    \item \textbf{Residual policy action regularizer}: This is an $\ell_2$ penalty that discourages modifications to the base policy that do not result in larger rewards. Specifically, we penalize the amount by which the residual action changes the gripper action in gripper width space: 
    $$r_{action} = -\|(1-a_r) * a_b\|_2$$
    \item \textbf{Task failure penalty}: A poor residual policy can cause the combined policy to entirely fail to solve the task, for example, by failing to grasp the berry, dropping it in the middle of transport, or causing the policy to continuously stall in midair. In these cases, the human operator manually assigns a large negative reward only at the last timestep of $r_{failure} = -10$. Otherwise, $r_{failure} = 0$. Note that the failure reward is not assigned when the robot applies too much force and squishes or breaks the berry; it is only assigned when the picking or transporting motions fail.
\end{itemize}

The total reward is $r = r_{force} + 0.01 *r_{action} + r_{failure}$.

Table~\ref{tab:rl_hyperparams} shows detailed hyperparameters for SAC.

\begin{table}[h]
    \centering
    \begin{tabular}{lr}
        \toprule
        \textbf{Hyperparameter} & \textbf{Value} \\
        \midrule
        Learning rate & 3e-4 \\
        Batch size & 256 \\
        (Polyak) parameter $\tau$ & 0.005 \\
        Discount factor $\gamma$ & 0.99 \\
        Num gradient steps per env step & 5 \\
        Entropy coefficient & learned \\
        Target network updates per env step & 1 \\
        Policy/critic architecture & MLP w/ hidden dimensions [256, 256] \\
        
        \bottomrule
    \end{tabular}
    \vspace{1em}
    \caption{Hyperparameters for real-world reinforcement learning with SAC. We largely keep to Stable Baselines3~\cite{stable-baselines3} defaults, and any unlisted hyperparameters take on their SB3 default values. One notable difference is that we use $5$ gradient steps per environment step as opposed to the default of $1$, which we found to improve sample efficiency.}
    \label{tab:rl_hyperparams}
\end{table}

\subsubsection{Evaluation Details}
During evaluation, we evaluate each policy over $20$ trials.  To be consistent and minimize potential bias, we perform comparative evaluations of policies in one sitting and programmatically randomize the order in which the trial for each policy will be evaluated. This order is unknown to the experimental evaluator.

For Table~\ref{tab:berry_residual_results}, we use strict criteria to determine if a berry remains ``intact'': A berry with \textbf{any visible tear or cut whatsoever} is considered \textbf{not intact}. Our residual policy keeps $60\%$ of the berries fully intact, but visual inspection shows that even those it does not keep intact are considerably less damaged than the ones handled by the baselines (Figure~\ref{fig:berry_rl_results}).

To compute the pressure value comparisons in Table~\ref{tab:berry_residual_results}, we first filter out any raw sensor values below 0.005 to select only potentially active taxels, then apply the mapping of sensor values to pressure values (kPa) obtained following the calibration procedure from Section~\ref{appendix: sensor calibration}, and finally compute the mean of the mapped pressure values over all timesteps in the $20$ evaluation trajectories in each setting.

\subsection{Details on Calibration and Model Transfer Experiments}

For this experiment, the task definition and setup are identical to that of the pen reorientation task in Section~\ref{appendix:task_setup}. We start by collecting a dataset of $50$ demonstrations for the task, in the same fashion as in Section~\ref{appendix:task_setup}. 

For the baseline experimental configuration, we verify that the policies achieve strong success rates with the sensors with which the training data was collected (source sensors). 

We then physically exchange the sensors that are attached to the left and right sides of the gripper to the right and left respectively (as our sensors are designed to be interchangeable), simulating a rearrangement of the sensors. Deploying the policy directly with this configuration is the ``Swapped (no calib.)'' setting. 
We also calibrate the sensors using values computed as in Section~\ref{appendix: sensor calibration} to map all readings first into pressure values, and then into the space of source sensor readings. Note that this does not require retraining the policy. The setting with the same policy model, but providing remapped tactile values as inputs, is denoted ``Swapped (calib)''. We conduct blind, order-randomized evaluations between the two ``Swapped'' settings. Unfortunately, we cannot do so with ``Source sensors'' because running evaluation on that setting requires physically adjusting the hardware, making it impossible to evaluate blindly. 

We repeat this experiment with a set of physically different sensors that were fabricated using the same procedure, but in different batches. This simulates a replacement of the sensors, or deployment on an entirely different robotic setup. These experiments are denoted ``Replaced (no calib.)'' and ``Replaced (calib.)''.    

For the DIGIT comparison, we test the policies from Section~\ref{sec:pen_task} first by physically swapping the \textbf{gel cartridge} between left and right DIGIT sensors. The visual appearance of the gel can vary from cartridge to cartridge as a result of variations in manufacturing. This setting is denoted ``Swapped'' in Table~\ref{tab:policy_transfer_experiments}.
Replacement of gel cartridges is a common maintenance operation that must be performed as the surfaces of DIGIT sensors accumulate wear over time.
While the DIGIT policy in the target sensor setting (swapped gels) does not complete the task in any of the $20$ trials, it exhibits reasonable behaviors such as reaching for and often grasping the pen. However, it shows a variety of failure modes such as mispositioning the gripper before attempting a grasp, and being unable to push the pen against the cabinet in an appropriate way to reorient it. It is likely that the differences in the visual appearances of the gels led to these errors that ultimately cause task failure.

One technique for reducing a policy's sensitivity to visual appearance is to train policies computed on difference images. We tested a setting where the policy's tactile input is a difference between the current tactile image observation and the image from the first step in the trajectory, which is assumed to be captured while the sensor is not deformed. This yields significantly improved performance in the swapped sensor setting (``Swapped (diff. image)'' in Table~\ref{tab:policy_transfer_experiments}).

We finally test the performance of the DIGIT difference image policies on entirely physically distinct sensors ordered from Gelsight Inc. (different housing, image sensor, gel) to simulate sensor replacement, denoted (``Replaced (diff. image)'').

\end{document}